\documentclass[referee, sn-mathphys, pdflatex]{sn-jnl}
\usepackage[utf8]{inputenc}
\usepackage{amsmath}
\usepackage{graphicx}
\usepackage{physics}
\usepackage{geometry}
\usepackage[
style = nature,
backend=biber,
doi=false,isbn=false,url=false,eprint=false,date=year,
sorting = none
]{biblatex}
\usepackage{amsmath}
\usepackage{amsthm}
\usepackage{amssymb}
\usepackage{mathtools}
\usepackage{physics}
\usepackage{bm}
\usepackage{floatrow}
\usepackage{subfig}
\usepackage{float}

\newcommand*{\dsum}{\displaystyle\sum\limits}
\newcommand*{\av}[1]{\left\langle #1 \right\rangle}
\newcommand*{\wmat}[1]{\mat{W}^{(#1)}} 
\newcommand*{\bvec}[1]{\vec{b}^{(#1)}} 
\newcommand*{\w}[2]{w_{#1}^{(#2)}} 
\newcommand*{\deltafc}[1]{\delta\left({#1}\right)} 
\newcommand*{\thetafc}[1]{\theta\left({#1}\right)} 
\newcommand*{\preact}[2]{z_{\mathcal{P},\vec{X}^{(m)}}({#1}, {#2})} 
\newcommand*{\U}[2]{\mathcal{U}_{#1}^{#2}}
\newcommand*{\yerr}{\Delta Y^{(m)}}

\renewcommand{\vec}{\boldsymbol}

\newcommand*{\dfn}[1]{\emph{#1}}

\newcommand*{\mathtext}[1]{\mathrm{#1}}

\newcommand*{\mpt}{\,.} 
\newcommand*{\mcm}{\,,} 

\newcommand*{\numberthis}{\addtocounter{equation}{1}\tag{\theequation}}

\newcommand*{\reals}{\mathbb{R}}

\newcommand*{\set}[1]{\left\{#1\right\}}

\newcommand*{\mat}[1]{\underline{\underline{\mathbf{#1}}}} 
\renewcommand*{\vec}[1]{\mathbf{#1}}
\newcommand*{\trp}[1]{\left[#1\right]^\mathrm{T}} 

\newcommand*{\actfc}[1]{%
  \ifx\relax#1\relax
    \rho 
  \else
    \rho\left(#1\right) 
  \fi
}

\title{Emergent weight morphologies in deep neural networks}
\author[1]{Pascal de Jong}
\equalcont{These authors contributed equally as first authors.}
\author[1]{Felix J. Meigel}
\equalcont{These authors contributed equally as first authors.}
\author*[1]{Steffen Rulands}
\email{rulands@lmu.de}
\affil[1]{Ludwig-Maximilians-Universit\"at M\"unchen, Arnold-Sommerfeld-Center for Theoretical
Physics, Theresienstr. 37, 80333 M\"unchen, Germany}
\date{\today}

\abstract{
Whether deep neural networks can exhibit emergent behaviour is not only relevant to understanding how deep learning works, but also pivotal for assessing the potential security risks of increasingly capable artificial intelligence systems. Here, we show that training deep neural networks gives rise to emergent weight morphologies independent of the training data. Specifically, using an approach akin to condensed matter physics, we derive from first principles a theory predicting that the homogeneous state of deep neural networks is unstable in a way that leads to the emergence of periodic channel structures. We verify these structures by performing numerical experiments on a variety of data sets. Our work demonstrates emergence in the training of deep neural networks, which impacts their achievable performance.
}

\begin{document}

\maketitle
\section*{Introduction}
Artificial intelligence is the imitation of human cognitive function by a computer. Recent breakthroughs in this field relied on the ability to train deep neural networks~\cite{hinton2006reducing} on large sets of data. These advances led to leaps in computer vision~\cite{voulodimos2018deep, szeliski2022computer}, natural language processing~\cite{goldberg2016primer,vaswani2017attention}, protein design~\cite{jumper2021highly,wang2018computational,omar2023protein} and others. In the simplest case, deep neural networks have a layered structure in which functional units, called neurons, are connected to neurons of neighbouring layers. The strengths of these connections are encoded in weights, which are determined by minimizing a cost function during training. 

Large neural networks have the capability of making generalizable predictions despite operating in an overparameterised regime~\cite{zhang2021understanding}. The effectiveness of deep neural networks has been explained by theoretical work based, for example, on analogies to information compression~\cite{shwartz2017opening, tishby2015deep}, energy landscapes in disordered systems~\cite{choromanska2015loss, geiger2019jamming,krauth1988roles,baity2018comparing, geiger2019jamming}, and statistical physics~\cite{geiger2021landscape,mezard2009constraint, geiger2020scaling,goldt2020modeling,d2020double,mehta2014exact,carleo2019machine}.
In light of the potential security risks of artificial intelligence~\cite{bender2021dangers,bentley2018should,bommasani2021opportunities}, the increasing capabilities of deep neural networks have raised the question of whether they can exhibit behaviour that does not originate from the training data. In the terminology of physics, this behaviour of deep neural networks is reminiscent of emergent phenomena, in which large-scale properties of complex systems go beyond the properties of the interactions between their components~\cite{schmelzer2013thermodynamics, anderson1972more}.

Empirical studies have indeed shown signs of this. For example, neural networks can abruptly gain new capabilities with an increasing number of parameters~\cite{wei2022emergent, ganguli2022predictability} or training time~\cite{power2022grokking}. For large language models, these abilities have been suggested to go beyond the scope of textual training data~\cite{bubeck2023sparks,brown2020language}. Models have recently been brought forward that explain emergence in artificial intelligence systems in terms of physical concepts like effective theories~\cite{liu2022towards, halverson2021neural}, superpositions~\cite{elhage2022toy}, broken power laws~\cite{caballero2022broken}, quantization~\cite{michaud2024quantization}, and phase transitions~\cite{achille2018emergence}. Because existing approaches do not directly link macroscopic phenomena to the microscopic training dynamics of deep neural networks, it remains a point of discussion whether the observations of the abrupt learning of new capabilities are a direct consequence of emergence~\cite{schaeffer2024emergent,lu2023emergent}.

To understand whether deep neural networks can exhibit emergent behaviour, we here follow a bottom-up approach that derives emergent properties from first principles. Starting from the transparent rules of weight updates during training, we employ a condensed matter approach to derive a theory of emergent, macroscopic structures in deep neural networks. Specifically, we show that emergent morphologies of weights in deep neural networks arise during their training. To this end, we treat neural networks as many-particle systems comprising interacting units that describe the local weight morphology. We derive the interactions between these units, and show that on the macroscopic level they give rise to channel-like structures that oscillate in width. Mathematically, this means that the homogeneous state of deep feedforward neural networks exhibits a morphological instability. Finally we show that these structures can have implications for the function and achievable performance of deep neural networks.

\section*{Results}
\begin{figure}
    \centering
    \includegraphics[width=1.0\linewidth]{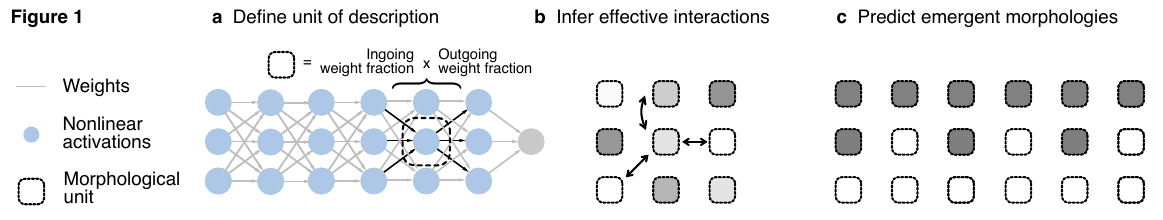}
    \caption{Illustration of the theoretical approach. \textbf{a} As a first step, we define a unit describing the local morphology of weights (dashed rectangle). This unit is mathematically represented by the product of in- and outgoing weight fractions of a node. \textbf{b} We then infer effective interactions between these morphological units, represented by the arrows in this figure. The shading represents the value of the morphological unit. \textbf{c} We finally predict emergent, large-scale morphological structures from these interactions. Shading as in b.}
    \label{fig:1}
\end{figure}

Neural networks of different architectures, like transformers and convolutional neural networks, all comprise non-linear nodes and linear connections between them. The strengths of these connections are termed weights. Independent of the specific architecture, neural networks are hierarchically organised into separate connected layers with multiple, mutually unconnected nodes in the same layer. During training, weights evolve to minimise a loss function on a given data set. Here we ask if this process gives rise to the emergence of large-scale order in the weight distribution, independently of the data. To investigate this, we start from the initial state of a neural network before training, in which weights take random values with low variance. We then ask whether this state becomes intrinsically unstable during training, in that any small perturbation gives rise to emergent, large-scale weight morphologies. To this end, we treat deep neural networks as a form of complex matter and take an approach akin to condensed matter physics: we first define the fundamental units that describe locally the weight morphology (Fig.~\ref{fig:1}a), then derive effective interactions between these units (Fig.~\ref{fig:1}b), and finally investigate the consequences of these interactions on the macroscopic scale (Fig.~\ref{fig:1}c).

\subsection*{Morphological description of deep neural networks}
Considering the layered structure of deep neural networks, the local weight morphology around a given node of the deep neural network is naturally described by how much a given node is connected to the previous and next layers. These quantities are mathematically represented by the ratios $\Omega_{\mathrm{in}}(n,l)$ and $\Omega_{\mathrm{out}}(n,l)$ between the sum of absolute weights going either in or out of a given node $n$ in layer $l$ and the total absolute weight between the respective layers. The nodal connectivity $r_n^{(l)}$ then is the product of both fractions (Fig.~\ref{fig:1}a),
\begin{equation}\label{eq:connectivity}
    r_n^{(l)} = \Omega_{\mathrm{in}}(n,l) \cdot \Omega_{\mathrm{out}}(n,l)\, ,
\end{equation}
which is bounded between 0 and 1 due to the normalisation of the weight fractions. We now aim to derive effective interactions between nodal connectivities. To this end, we first quantify the coevolution of pairs of weights, from which we then derive effective interactions between nodes. For the time evolution of weights, we use stochastic gradient descent with learning rate $\eta$ in a potential given by the loss function $\mathcal{L}$,
\begin{equation}
    w\longleftarrow w - \eta\pdv{\mathcal{L}}{w}\, .
\end{equation}
The loss function quantifies the deviation of the network prediction from the true data labels, and we assume the common choice of the squared-error loss function. We then represent the output of a neural network with one output node and a fixed number of nodes $N$ in each hidden layer as a function of all possible paths between the input and output layer~\cite{choromanska2015loss}. For any pair of weights in adjacent layers sharing a common node, we obtain an exact expression for the coevolution of their values, $w$ and $w'$, and their respective increments after one step of training, $\Delta w$ and $\Delta w'$. We find that weights in adjacent layers are positively coupled, and this coupling is $N$ times stronger than for weights in nonadjacent layers (Supplementary Theory).

\begin{figure}
    \centering
    \includegraphics[width=1.0\linewidth]{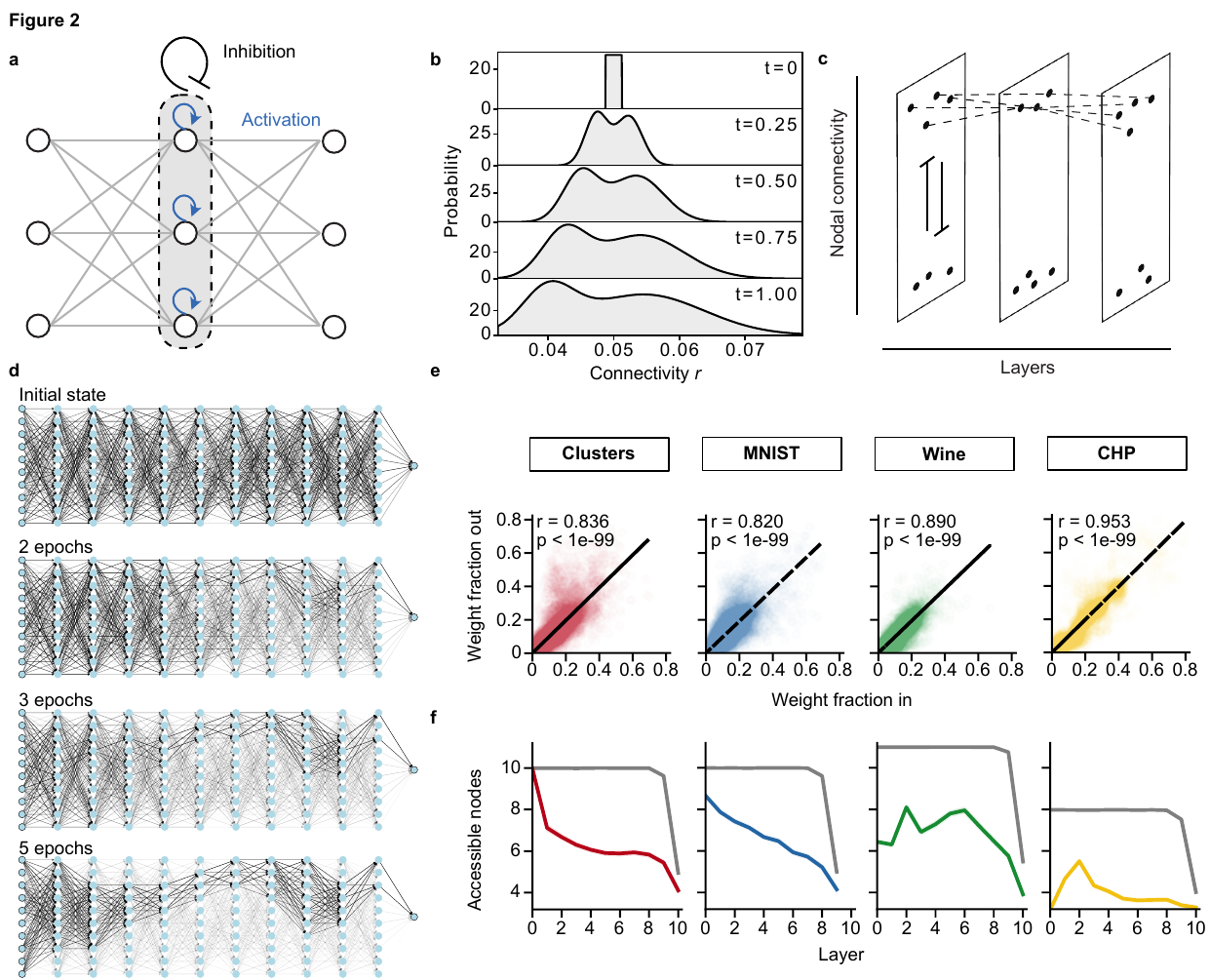}
    \caption{\textbf{a} Schematic depicting effective interactions between nodes in the same layer. \textbf{b} Numerical solution of Eq.~\eqref{eq:intralayer} using a 5th order Runge-Kutta scheme. The simulation uses 20 nodes in the layer and a uniform initial distribution for both the connectivities $r_i$ and constants $c_i$. Each simulation ran for 250 timesteps ($t_{\text{max}}=1.0$), and 1000 simulations were aggregated. \textbf{c} Schematic showing the mechanism leading to channel formation. \textbf{d} Snapshots of a neural network at different stages during early training. The shade of the connecting lines denotes the relative absolute strength of the weight with respect to the maximum within each layer. The neural network was trained on synthetic cluster data. Nodes in all images are ordered from top to bottom by the value of their connectivity after training. \textbf{e} Outgoing weight fraction $\Omega_{\text{out}}$ as a function of the ingoing weight fraction $\Omega_\text{in}$ for three different data sets. Each point corresponds to an individual node from 250 networks in total. Pearson correlation coefficient $r$ and $p$-value are shown. Dashed lines correspond to a linear fit through the origin. \textbf{f} Number of accessible nodes when traversing the network backwards from output to input, after pruning away all weights smaller in absolute value than the mean. Grey lines correspond to values computed before training, colored lines denote values computed after training. Layer difference 0 corresponds to the input layer. Shaded areas denote standard errors.}
    \label{fig:2}
\end{figure}

\subsection*{Channel morphologies}
This framework then allowed us to derive effective time-evolution equations for the nodal connectivities during training in fully connected feedforward neural networks under the assumption of an initial weight distribution with a small variance and mean. In this case, weights will initially grow in absolute strength during training (Supplementary Theory). To the highest order in the fluctuations around the homogeneous state, the time-evolution of nodal connectivities is dominated by effective interactions between neurons in the same layer,
\begin{equation}\label{eq:intralayer}
    \dv{r_j^{(l)}}{t}\approx c_jr_j^{(l)}\left(1-\sqrt{r_j^{(l)}}\right) - r_j^{(l)}\sum_{i\neq j}{c_i\sqrt{r_i^{(l)}}} \,.
\end{equation}
Here, the prefactors $c_i$ capture all higher-order effects from interactions with neurons in adjacent layers. Very close to the homogeneous state, these prefactors are constant in time and equal across all neurons in the same layer, and they are positive under the assumption of growing weights above. The first term in Eq.~\eqref{eq:intralayer} shows that during training, connectivities undergo bounded growth with a rate given by $c_j$. The second term describes effective repressive interactions with all other neurons in the same layer~\cite{patalano2022self} (Fig.~\ref{fig:2}a).

In the homogeneous state, all $c_j$ and $r_j^{(l)}$ take the same value and the two terms in Eq.~\eqref{eq:intralayer}  cancel each other out (Supplementary Theory). The homogeneous state is therefore a fixed point of Eq~\eqref{eq:intralayer}. Any perturbation of the homogeneous state leads to $c_j$ and $r_j^{(l)}$ taking values that differ between nodes. Then, a given nodal connectivity $r_j^{(l)}$ will grow if $c_j>\av{c_i}_{r_i^{(l)}}$, where the average is taken with respect to the distribution of nodal connectivities, and shrink otherwise. Close to the homogeneous state, this distribution is uniform. Because the growth and shrinkage of connectivities are bounded, these dynamics give rise to a bimodal distribution of connectivities in each layer, which is corroborated by numerical integration of Eq.~\eqref{eq:intralayer} (Fig.~\ref{fig:2}b).

By the definition of the connectivities, Eq.~\eqref{eq:connectivity}, 
nodes that are strongly connected, which we refer to as upper mode nodes, are also strongly connected to nodes in the upper mode of adjacent layers (Fig.~\ref{fig:2}c). Vice versa, nodes in the lower mode of the distribution are only weakly connected between layers. On the scale of the entire neural network, this is predicted to give rise to the formation of channel-like morphologies between the input and the output layer.

To test these predictions empirically, we trained a large number of neural networks on a variety of benchmark data sets (Methods): a synthetic cluster dataset, the MNIST classification dataset of handwritten digits~\cite{lecun1998mnist}, the white wine quality dataset~\cite{wine_quality_186}, and the California Housing regression dataset~\cite{pace1997sparse}. To facilitate direct comparison with theoretical predictions, we used networks with a constant number of nodes per layer, a single node in the output layer, and ReLU activations. We trained deep neural networks on these data sets using mini-batch gradient descent and recorded at each training episode the value of each weight. Figure~\ref{fig:2}d shows a visual representation of the relative absolute strength of the weights throughout early times of training (epoch 0 to 5 of a total 250) in an exemplary training run. It visually confirms the formation of a channel structure during the early episodes of training. The formation of channel-like structures is also confirmed by a statistical analysis of the Pearson correlation $r$ between the strength of ingoing and outgoing weight fractions, $\Omega_{\text{in}}$ and $\Omega_{\text{out}}$, that define the connectivities of individual nodes (Fig. \ref{fig:2}e). Furthermore, Fig. \ref{fig:2}f shows a network analysis of the number of accessible nodes in each layer, after pruning away all weights smaller than the mean (Methods). In a network with random weights, the accessibility remains constant, whereas in the trained network it decreases significantly, indicating that the nodes with large correlated ingoing and outgoing fractions are focused on a subset of nodes in each layer, and these subsets are connected.

\subsection*{Periodic channel amplitudes} 
\begin{figure}
    \centering
    \includegraphics[width=1.0\linewidth]{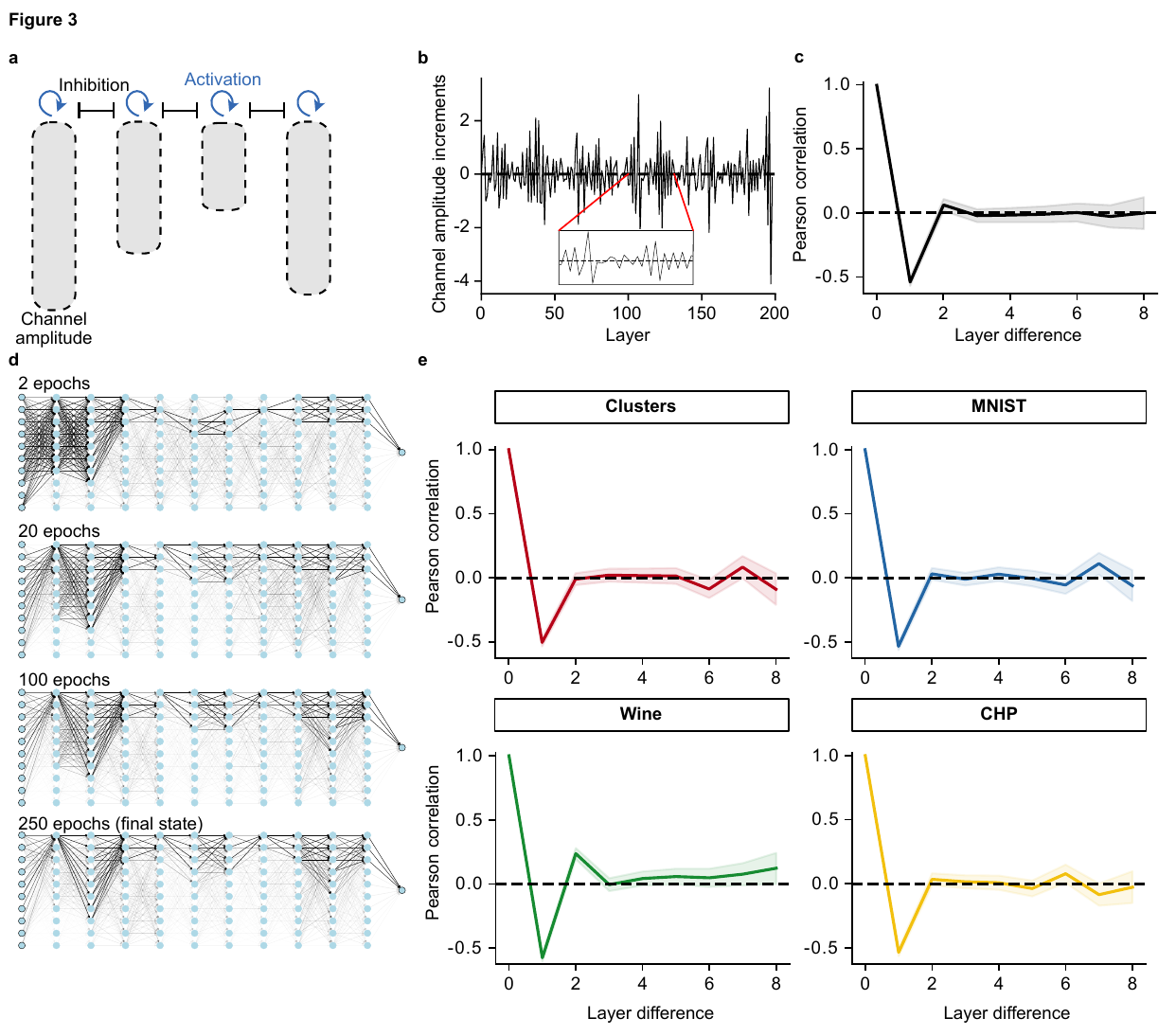}
    \caption{ 
    \textbf{a} Schematic illustrating lateral interactions modulating the channel amplitude. \textbf{b} Amplitude increments between consecutive layers obtained from numerical solution of the connectivity dynamics with coupling to neighbouring layers using a 5th order Runge-Kutta scheme. The simulated network consisted of 10 nodes and 200 layers, and both initial connectivities and constants were drawn from a uniform distribution. The inlay shows an exemplary region with oscillating, anticorrelated amplitude increments. \textbf{c} Pearson autocorrelation of numerical simulations as in \textbf{b}. Each simulation consisted of 10 nodes and 12 layers, and both initial connectivities and constants are drawn from a uniform distribution. In total 250 simulations were aggregated. The shaded area denotes the standard error. \textbf{d} Snapshots of a neural network at different stages after channel formation until the end of training. The neural networks was trained on synthetic cluster data. Line shades and nodal permutation as in Fig.~\ref{fig:2}d. \textbf{e} Pearson autocorrelation of amplitude increments as a function of the layer difference. Shaded areas denote standard errors.}
    \label{fig:3}
\end{figure}

At later times during training, when the variance in the connectivities in adjacent layers has increased, the nonlinear dependence of the coupling terms $c_j$ in Eq.~\eqref{eq:intralayer} on these adjacent layers becomes important. We, therefore, asked whether the channel structure that arises due to the instability at short times gets modulated due to the higher-order coupling between layers at later stages during training. To investigate this, we defined an amplitude variable $a_l\equiv N\sum_j r_j^{(l)}$. In the homogeneous state, each of the $N$ connectivities takes a value $N^{-2}$, such that in this case $a=1$ and the channel is wide. If all the connectivity is focused onto a single node with $r_j^{(l)}=1$ and 0 for all others, we see that $a=N$ and the channel has minimum width. 
By explicitly considering the nearest neighbour layer dependencies of the layer-couplings $c_i$ in Eq.~\eqref{eq:intralayer} we derive coupled differential equations for the amplitudes $a_l$ (Supplementary Theory). To the highest order in the fluctuations of individual connectivities $r$ this time evolution of $a_l$ comprises an interaction term of the form,
\begin{equation}
    a_{l}\left(1-\sqrt{a_{l}}\right)\left(c^{\text{R}}\sqrt{a_{l+1}} +c^{\text{L}}\sqrt{a_{l-1}} \right) \, , \label{eq:channel-width}
\end{equation}
where, $c^{\text{R}}$ and $c^{\text{L}}$ summarise higher order, non-nearest neighbour contributions from the right and left layers. Because of the bounds on $a_l$, $1-\sqrt{a_{l}}$ is non-positive, such that this term always leads to a decrease in the value of $a_l$, and this decrease is directly coupled to the amplitudes $a_{l\pm 1}$ in adjacent layers. This interaction term therefore represents an inhibition by the channel amplitudes in neighbouring layers (Fig.~\ref{fig:3}a). Equation~\eqref{eq:channel-width} thus gives rise to a local anticorrelation of the channel amplitudes in adjacent layers. Globally, this yields an oscillatory modulation of the channel amplitude (Fig.~\ref{fig:3}b). For deep neural networks of a finite size this oscillatory modulation is influenced by the boundaries defined by the input and output layers. This is reflected in a decrease of the correlation function of channel amplitudes. Figure~\ref{fig:3}c shows the correlation function for neural networks of the depth as was used for our numerical experiments.

To test these predictions empirically we analyzed weight morphologies of deep neural networks throughout later stages of training. Figure~\ref{fig:3}d shows exemplary representations of the neural network morphology after the initial formation of a channel morphology up to the end of training. Figure \ref{fig:3}e shows for different training tasks that after training, the changes in the channel amplitude become anticorrelated between consecutive layer. This reflects a periodic modulation of the channel amplitude with a periodicity of two layers.

\subsection*{Perturbations}
The results above show that the homogeneous state of deep neural networks admits a self-organised instability, which gives rise to complex weight morphologies. Our theory also predicts in which cases this instability does not occur. This is the case if the initial state of the deep neural network is homogeneous but has a large mean, as well as when it is not homogeneous and the weights have a high variance.

To illustrate this, we trained networks with initially uniformly distributed weights and varying standard deviations. For each value of the standard deviation, we computed the Pearson correlation between the in- and outgoing weight fraction, as in Fig.~\ref{fig:2}e and the number of accessible nodes as in Fig.~\ref{fig:2}f. With increasing standard deviation of the initial weight distribution, the correlation between in- and out-going weight fractions decreases (Fig.~\ref{fig:4}a) and the number of accessible nodes increases (Fig.~\ref{fig:4}b). This indicates that channel formation breaks down, in support of our prediction that pattern formation does not occur for large initial variances.

\subsection*{Implications for performance}
\begin{figure}
    \centering
    \includegraphics[width=1.0\linewidth]{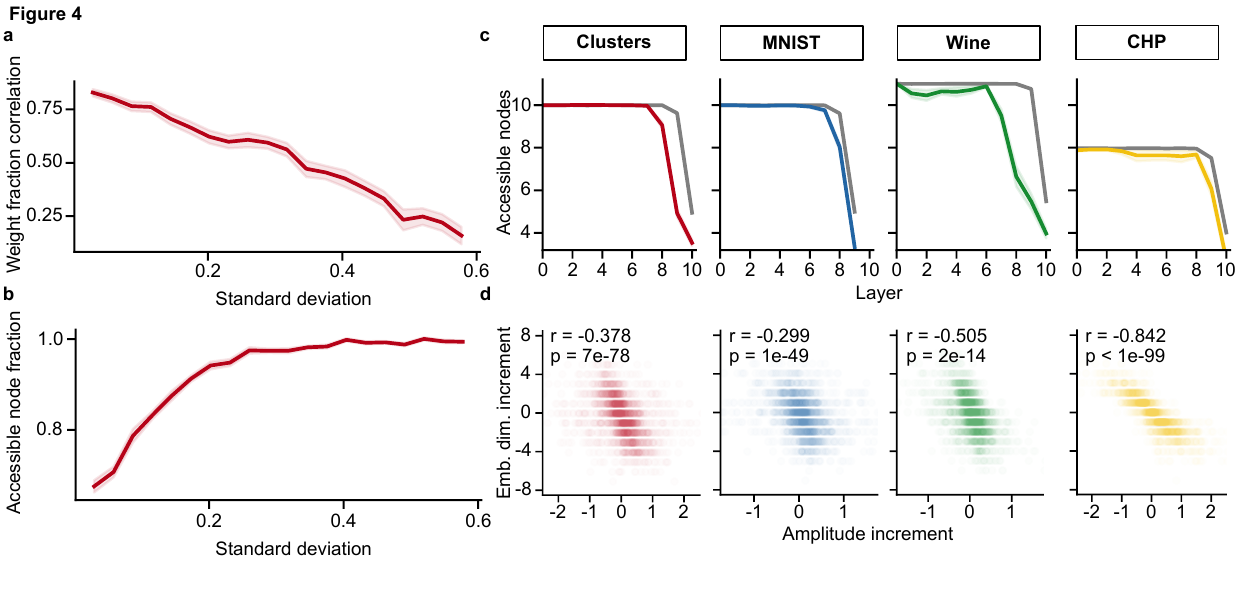}
    \caption{\textbf{a} Pearson correlation of in- and outgoing weight fractions as in Fig.~\ref{fig:2}e as a function of the standard deviation of the initial weight distribution. \textbf{b} Fraction of accessible nodes, computed as the ratio between the colored line and the gray line in Fig.~\ref{fig:2}f, as a function of the standard deviation of the initial weight distribution. \textbf{c} Number of accessible nodes for poorly trained networks with an accuracy below 20\% before (gray line) and after (colored line) training, after pruning away all weights smaller in absolute value than the mean. Shaded areas denote standard errors. \textbf{d} Amplitude increments from one layer to the next scattered against the corresponding change in embedding dimension. Each point thus corresponds to a comparison of two successive layers. Pearson correlation coefficient $r$ and $p$-value are shown.}
    \label{fig:4}
\end{figure}

So far, we have shown that neural networks exhibit an instability that gives rise to emergent weight morphologies. Although this instability is independent of the training data, the question arises whether these structures carry significance for the performance of the network. Neural networks trained in conditions where structure formation was predicted to fail still showed high accuracy in some cases, implying that structure formation is not a strictly necessary condition for optimal training in these cases.
However, optimal training typically implies structure formation if the initial conditions are such that it can occur. To illustrate this, Fig.~\ref{fig:4}c shows the number of accessible nodes per layer in the poorly trained networks, with an accuracy below 20\%, for each of the four datasets. Because the curves corresponding to trained and untrained networks now overlap, we can no longer identify a channel of strongly connected nodes. Poor training thus corresponds to a lack of channel formation in these cases. While the accuracy achieved is highly dependent on hyperparameters, this indicates that the formation of self-organised weight morphologies might contribute to the function of deep neural networks.

To study this function, we asked if the dimensionality of the data representations in the hidden layers of the network varies in the same manner as the oscillatory weight pattern. To test this,
we quantified the embedding dimension of the hidden data representation~\cite{ansuini2019intrinsic} in individual layers of deep neural networks before and after training. We did this by computing the largest number of nodes in each layer that were active in at least one data sample. Figure~\ref{fig:4}d shows that increments of the embedding dimension correlate with increments of the channel amplitude. This implies that hidden data representations oscillate in the same manner as the channel width. This is non-trivial because, while the network morphology is a property solely of the weights, the embedding dimension is highly dependent on the nonlinear activations and only indirectly related to the weights.

These results imply that the number of nodes that is used to represent the data, the embedding dimension, varies periodically throughout the layers. Such dimensional changes in the data representation are ubiquitously used in machine learning algorithms to make predictions on complex data~\cite{berahmand2024autoencoders,perez2004kernel}. Increasing the dimensionality of data representations leads, according to Cover's theorem~\cite{cover1965geometrical}, to a high probability of linear separability of complex data structures. Such dimensionality transformations are used in the kernel method and feature engineering. Vice versa, reducing the dimensionality of data representations leads to compression, which has been shown to aid learning by facilitating generalization~\cite{shwartz2017opening}. The observed correlation between the embedding dimension of hidden data representations and the emergent oscillatory weight structures indicates that these structures might also facilitate the repeated transformation of data representations to higher dimensions, as in the kernel method~\cite{kerneltrick}, and back to lower dimensions, as in autoencoders~\cite{berahmand2024autoencoders,kingma2013auto}. This connection suggests that the oscillating weight morphology can potentially be used to improve the function of deep neural networks. In general, emergent weight morphologies provide the foundation on which learning occurs and may both facilitate and constrain deep learning.

\section*{Discussion}
We showed that deep neural networks exhibit emergent behaviour during training. Specifically, the homogeneous state, in which the weights take random values with low variance, exhibits an instability which gives rise to complex weight morphologies independent of the training data. In the early stages of training, this leads to the formation of a channel structure of highly connected weights, which then, during later training times, is periodically modulated in amplitude.

We derived these results for the specific case of fully connected feedforward neural networks with ReLU activation functions, but they extend to all neural networks with a feedforward architecture whose output can be expressed as a sum over all paths through the neural network. These include convolutional neural networks which can be mapped to sparse deep neural networks. Recent work has shown that an analogous path framework also exists for transformers~\cite{elhage2021mathematical}. Our results are specific to training algorithms based on gradient descent with a squared-error loss function. They are however not limited to ReLU nonlinearities but can be applied to neural networks with general sigmoidal activation functions as long as they can be approximated by a piecewise linear function (Supplementary Theory).

The resulting structures emerge independently from the training data and are therefore not necessarily involved in the function of the neural network. However, they do impose universal morphological constraints under which the network learns to make generalizable predictions. Beyond constraining the learning dynamics, these structures may also benefit learning. We showed that there are correlations between these structures and the dimensionality of the data embedding in the neural network. Oscillating embedding dimensions do not necessarily lead to better learning, but if they are combined with appropriate nonlinear data transformations, they may aid in the detection of important data features. As an example, transformations in the embedding dimension are used in the kernel trick for classification as well as in autoencoders.

Furthermore, the lottery ticket hypothesis posits that in dense neural networks there exist sparse subnetworks which can achieve comparable performance to the entire network \cite{frankle2018lottery}. It has been suggested that the random initialisation of these subnetworks makes them very well suited for learning.
Here, we have shown that such sparse networks can emerge via self-organization from the training dynamics independently of the training goal. This raises the question if the emergent weight morphologies correspond to the sub-networks that the lottery ticket hypothesis identifies as critical to efficient learning. Pruning methods commonly used to isolate ``winning tickets'' could be improved by making use of the self-organization principles we described here.

Finally, the question of whether artificial intelligence systems exhibit emergent behaviour is relevant to the discussion of the security of large artificial intelligence systems, as emergent behaviour may lead to unpredictable capabilities. Our work shows that emergence already exists in relatively simple neural networks, and this also influences learning. It raises the question of whether emergent structures lead to entirely new capabilities in more complex architectures.

\section*{Methods}
\subsection*{Data sets used for numerical experiments}
In this research, four datasets were used. A synthetic cluster dataset, where 10,240 points were organised in 11 clusters each with a Gaussian distribution with a fixed standard deviation of 0.05, and where each point has 10 positional coordinates, which function as the input features for training. Each cluster is labeled 1 to 11 and the neural network was trained to predict this class. Out of all the samples, 8192 were used for training, and 2048 for testing.
Secondly, the MNIST dataset of handwritten digits consisting of 70,000 images. The images were flattened such that each pixel corresponds to a single input feature, and the network was trained to predict the written digit from these pixels. 60,000 images were used for training, and 10,000 for testing.
Thirdly, the white wine quality dataset, which contains 4898 samples with 11 features each. Here 3918 samples were used for training and 980 for testing.
Finally the California Housing Price (CHP) dataset, which contains 8 features for a total of 20,640 samples, out of which 16,512 were used for training and the rest for testing.

\subsection*{Training of deep neural networks}
For the building and training of neural networks, the open-source \verb|Python| (version 3.12) library \verb|keras|~\cite{chollet2015keras} (version 3.7) was used. The code produces fully connected deep neural networks, with initial weights and biases drawn from a uniform distribution, $U\left(-0.05,0.05\right)$.

For the synthetic and wine datasets, the number of nodes in each layer was constant and set to be equal to the number of input features in the data. For the synthetic case, this corresponds to 10 nodes, and for the CH data, there were 8 nodes. In both cases, 10 hidden layers were added. For MNIST, the number of input features is the number of pixels in the images, 750, so in this case we decided to add 2 intermediate layers with 128 and 32 nodes respectively, before scaling down to 10 hidden layers with 10 nodes, in which the structure formation was studied. We always considered a single linear output node, and all hidden nodes apply a $\text{ReLU}$ activation function, in line with our theoretical framework.

For training, a mean-squared error loss function was used, together with the Adam optimiser~\cite{kingma2014adam}, and an initial learning rate of 0.01. Training of 500 networks for each dataset was carried out in mini-batch sizes of 256, for a total of 250 (synthetic, wine, CHP) and 150 (MNIST) epochs. After training the test accuracy was recorded, and unless stated otherwise, only the networks with an accuracy larger than the median of all networks were considered, to filter out those that failed to learn.

For the snapshots of neural networks during training, we used the synthetic dataset and recorded the weights after every 4 minibatches for a total of 250 epochs, where one epoch consists of 8 minibatches. 

For the sweep over variances of the uniform initial weight distribution, we trained 30 networks per standard deviation, which ranges from 0.03 to 0.6.

\subsection*{Calculation of the number of accessible nodes}
The plot in Figure 3 showing the number of accessible nodes as a function of layer, is obtained as follows. First, we consider only the absolute value of all weights and prune away all those that are below the mean value of weights. Due to this pruning, if all outgoing connections from a node in layer $l$ have been cut, any input to this node can no longer access layer $l+1$. In each layer, the number of accessible nodes from output back to input after the pruning is counted, and this is plotted as a function of layer depth, where layer difference 0 corresponds to the input layer.

\subsection*{Calculation of correlation functions}
Correlation functions shown in Figure~\ref{fig:3} are correlations of amplitude increments. This means, that after computing the amplitude of each layer in a network, the 1 layer differences were extracted, and these differences are used to compute the correlation function. The reason for this is that it is not specific values of the amplitude that are expected to be correlated, but changes in the amplitude. For example, the negative correlation at a difference of 1, means that the increment from layer $l-1$ to layer $l$ typically has the opposite sign as the increment from layer $l$ to layer $l+1$. It does not necessarily imply that the value of the amplitude in layer $l$ is negatively correlated with the value of the amplitude in layer $l+1$, which would be the correlation of amplitudes, as opposed to their increments.

\subsection*{Calculation of the embedding dimension}
In principle, the number of nodes in each layer is fixed, and cannot change. We therefore instead define a data-based embedding dimension which quantifies the size of the external space the data manifold lies in. We know that each node applies a ReLU activation to its input, outputting either 0 or a positive number. Therefore, when the output of a node is zero for a certain input instance, we can discard this node and only consider the information coming from the active nodes. Although the hidden representation of a data instance in layer $l$ has $n_l$ coordinates, we discard all the dimensions of this space for which the activation state is 0. This leads to an effective reduction of the space of the hidden data representation, and we define this embedding dimension of a layer as
\begin{align}
    d_{\text{ED}}(l) &\equiv \max_{m}\{N_{\text{active}}(l)\} \\
\end{align}
This is the maximum number of active nodes in a layer across all instances $m$ of the data. This definition does not keep track of which dimensions we discard for each sample, but due to the nodal permutation symmetry, assigning a fixed label to each dimension and keeping track of this as well is somewhat arbitrary.

\section*{Acknowledgements}
We thank Daniel Pals and the entire Rulands group for critical discussions.
This project has received funding from the European Research Council (ERC, grant agreement no. 950349).


\section*{Code availability}
Simulation routines are described in the methods section. Code snippets are available from the corresponding author upon reasonable request.

\printbibliography

\pagebreak
\begin{center}
\textbf{\large Supplementary Theory to 'Emergent weight morphologies in deep neural networks'}
\end{center}
\setcounter{equation}{0}
\setcounter{figure}{0}
\setcounter{table}{0}
\setcounter{page}{1}
\makeatletter
\renewcommand{\theequation}{S\arabic{equation}}
\renewcommand{\thefigure}{S\arabic{figure}}

The structure of this supplement is as follows. We first introduce the path and activity framework as originally proposed in Ref.~\cite{choromanska2015loss}. We then build on this and derive the coupled time evolution of weights in this framework. We then coarse grain to connectivities, and derive analogous equations on this level of description. Finally, we study how these equations give rise to patterns.

\section{Active path representation of deep neural networks}
Before introducing our framework, we can already list some of the criteria it should fulfill, simultaneously motivating our choice. First of all, we would like to view a neural network as an object with a spatial extent, such that our physical intuition about what it means to have structure formation in a system is justified. This perspective should therefore be inherent to the framework, allowing for a straightforward, physically intuitive interpretation of any results we obtain from it. The second criterion stems from the fact that we want to study structure formation in the weights. However, in the usual definition weights and nonlinearities can not be seen separately, as there is always a node in between weights where an activation function is applied. Ideally, our formalism should single out the weights, and to some extent decouple them from the nonlinear activations, such that it actually makes sense to think about weight structure formation.

\subsection{Definition of the path-activity formalism}
To construct an analytic theory of structure formation in deep neural networks, we focus on the fully-connected feedforward setting. Let 
\begin{align}
    \mathcal{N}_{\alpha,\mathcal{P}}:\quad \reals^{d\times M} &\longrightarrow \reals^{n_H\times M} \\
    \mat{X} &\longmapsto \hat{\mat{Y}}
\end{align}
be such a network with architecture $A=(\set{n_i}_{i=1}^{H}, \actfc{})$ and parameters $\mathcal{P}=\left(\set{\wmat{l}}_{l=1}^{H}, \set{\bvec{l}}_{l=1}^{H}\right)$, that maps an input $\mat{X}\in\reals^{d\times M}$, containing $M$ samples with $d$ features, to an output $\hat{\mat{Y}}\in\reals^{n_H\times M}$. The mapping $\mat{X} \longmapsto \hat{\mat{Y}}$ can be written as
\begin{equation}\label{eq:nested_NN_output}
    \hat{\mat{Y}} = \actfc{\trp{\wmat{H}}\actfc{\ldots\trp{\wmat{2}}\actfc{\trp{\wmat{1}}\mat{X}}\ldots}}\mcm
\end{equation}
for weight matrices $\wmat{i}\in\reals^{n_{i-1}\times n_{i}}$, and setting all biases to zero. However, under the assumption of a single, linear output node and the rectified linear unit as the activation function for the rest of the network, there exists an equivalent representation. This alternative stores all the nonlinear information about the network in a new object called the \dfn{activity} and expresses the output by linearly coupling the weights to this nonlinear object. To define this mathematically, we first introduce the concept of a \dfn{path}.

\subsubsection{Definition of paths}
Start by choosing one of the input features and label this by $i\in\set{1,\ldots,n_0=d}$. Next, pick one node in each subsequent layer, and give this set of nodes the label $j$, containing $H-1$ elements. The total number of such sets is given by
\begin{equation}
    \Gamma = \prod\limits_{i=1}^{H}{n_i}\mcm
\end{equation}
so that $j\in\set{1,\ldots,\Gamma}$. Each combination $(i,j)$ specifies a unique set of nodes $\set{n_{{i_j}^{(l)}}}_{l=0}^{H}$ obtained after choosing one node in each layer of the network\footnote{The choice for an input feature $i$ is regarded as equivalent to choosing an input node $i$.}. Now, for a given choice of $i$ and $j$, denote by $\w{i_j}{k}$ the weight that connects node ${i_j}^{(k-1)}$ to node ${i_j}^{(k)}$. The full set $\set{\w{i_j}{k}}_{k=1}^{H}$ then defines a unique connection from input $i$ to the output node, which we refer to as a \dfn{path} with label $i_j$. The set of all paths we name $G$, with size
\begin{equation}
    \abs{G}=n_0\cdot\Gamma\mpt
\end{equation}
Note that throughout this thesis we now employ the following notation for weights. The upper index always specifies the layer that a weight is connecting to, for example $\w{i_j}{k}$ connects two nodes in layers $k-1$ and $k$. The lower indices denote which nodes the weight connects, and here there are two possibilities. If the lower indices have a form as in $\w{i_j}{k}$, then this weight refers to the one that connects layer $k-1$ to $k$ along path $i_j$. Since multiple paths can use the same weight, this notation is not unique. If the lower indices have a form as in $\w{ab}{k}$, then this is the weight connecting nodes $a^{(k-1)}$ and $b^{(k)}$.

Consider a single input sample $\vec{X}^{(m)}\in\reals^{d}$, $m\in\set{1,\ldots,M}$. In the language introduced above, each individual component of this vector forms the starting point of multiple unique paths, and we denote these components by $X_{i_j}^{(m)}$, subject to the duplicate condition that
\begin{equation}
    X_{i_j}^{(m)}=X_{i_k}^{(m)} \; \forall\, k\in\set{1,\ldots,\Gamma}\mpt
\end{equation}
This merely reflects the fact that multiple unique paths originate from the same input feature. To continue from here, a second object is required, namely the activity of a path.

\subsubsection{Definition of nodal activities}
Paths as introduced above do not take the nonlinear nature of the network into account. Therefore, we construct the \dfn{path activity} to capture the effect of the activation functions applied in each node. In the case of the rectified linear unit, we can use the semi-linear property that any negative input is suppressed and leads to an output of zero, whereas a positive input passes through perfectly linearly without any modification, as it follows from the definition,
\begin{equation}
    \mathtext{ReLU}(x) = \max\set{0,x}\mpt
\end{equation}
Let us remind ourselves that in the biasless situation, the input of a node ${i_j}^{(l)}$ is the pre-activation value $\preact{i_j}{l}$ defined as
\begin{align}
    \preact{i_j}{l} &\equiv \left[\trp{\wmat{l}}\actfc{\ldots\actfc{\trp{\wmat{1}}\vec{X}^{(m)}}\ldots}\right]_{i_j^{(l)}}\mcm \\
    \preact{i_j}{1} &= \left[\trp{\wmat{1}}\vec{X}^{(m)}\right]_{i_j^{(1)}}\mpt
\end{align}
In general, this value depends on the specific input sample $\vec{X}^{(m)}$ and the full set of network parameters $\mathcal{P}$. The index ${i_j}^{(l)}$ denotes that from the vector
\begin{equation}
    \trp{\wmat{l}}\actfc{\ldots\actfc{\trp{\wmat{1}}\vec{X}^{(m)}}\ldots}
\end{equation}
with all pre-activations of layer $l$, we select the value for the node along the path under consideration. Using the Heaviside step function
\begin{equation}
    \thetafc{x}=
    \begin{cases}
    0,\;x < 0 \mcm \\
    1,\;x\geq 0 \mcm
    \end{cases}
\end{equation}
we define the activity $A_{{i_j}^{(l)}}^{(m)}\in\set{0,1}$ of a node as
\begin{equation}
    A_{{i_j}^{(l)}}^{(m)}(\vec{X}^{(m)}, \mathcal{P}) = \thetafc{\preact{i_j}{l}}\mcm
\end{equation}
expressing the semi-linear nature of $\mathrm{ReLU}$ as a binary number. If the input to a node is positive, the activation state is $1$, reflecting the fact that the input is passed on without modification. If on the other hand the input is negative, the node has an activity of $0$ and the input information does not propagate any further. Finally, we define the activity of a path $i_j$ as
\begin{align}
    A_{{i_j}}^{(m)}(\vec{X}^{(m)}, \mathcal{P}) &\equiv \prod_{l=1}^{H}{A_{{i_j}^{(l)}}^{(m)}(\vec{X}^{(m)}, \mathcal{P})} \\
    &= \prod_{l=1}^{H}{\thetafc{\preact{i_j}{l}}}\mpt \label{eq:path_activity}
\end{align}
This quantity again takes on a binary value of $1$ when all nodes along a path are active and $0$ if one or more are inactive.

\subsubsection{The path-activity output equation}
We now have all the ingredients and introduced all the notation to rewrite Eq. \eqref{eq:nested_NN_output} in terms of paths and activities. For an architecture as defined above, the map $\mathcal{N}_{\alpha,\mathcal{P}}:\quad \reals^d \longrightarrow \reals$ of a single input instance $\vec{X}^{(m)}\in\reals^d$ to an output $\hat{Y}^{(m)}\in\reals$ can be written as~\cite{choromanska2015loss}
\begin{equation}
    \hat{Y}^{(m)} = \dsum_{i=1}^{n_0}\dsum_{j=1}^{\Gamma}{X_{i_j}^{(m)}A_{i_j}^{(m)}\prod\limits_{k=1}^{H}{\w{i_j}{k}}}\mpt
\end{equation}
The first two summations run over all possible paths $i_j$ one can take through the network. The path activity $A_{i_j}^{(m)}$ acts as a delta-function, restricting the summation to the subset of active paths, since for inactive paths its value will be 0 and these terms therefore do not contribute to the output. Each active path contributes a factor given by the input feature $X_{i_j}^{(m)}$ of that path multiplied with the product $\prod\limits_{k=1}^{H}{\w{i_j}{k}}$ of all weights along it. For mathematical convenience we can combine the first two summations using the set of all paths $G$,
\begin{equation} \label{eq:path_activity_output}
    \hat{Y}^{(m)} = \dsum_{i_j\in G}{X_{i_j}^{(m)}A_{i_j}^{(m)}\prod\limits_{k=1}^{H}{\w{i_j}{k}}}\mpt
\end{equation}
Before delving into the application this formalism, let us emphasise its implications, and the reason why this formalism is a powerful tool for us to study structure formation in deep neural networks.

In a physical system, the existence of a structure typically implies that there is a space in which this structure resides. A neural network is a graph, which means that the notion of space is ill-defined, since we do not fix the embedding space it lives in. The notion of distance and spatial relationships is thus not intrinsic to the graph itself. It is for example not fully clear where one node should be positioned with respect to another, recalling the full permutation symmetry of all nodes in a layer. There is, however, one statement that always remains true, namely that information flows from layer to layer, from input to output. We can therefore naturally interpret this as the spatial dimension we would look for in a physical system. We then recognise that it is precisely this dimension that is also captured by the concept of a path as defined above. In other words, we could interpret the path formalism as a way of defining the internal spatial structure of a network. This shifts our picture of a deep neural network to one where the internal structure is not merely a sequence of layers, but rather a complex web of paths interwoven within the neural architecture. In that case, the very general idea of studying structure formation reduces to the well-defined problem of studying path statistics within this complex web. Through the lens of the path-activity formalism  we have thus not only defined what it means for a network to have an internal spatial structure, but we have also found a natural way of studying it through path statistics.

\subsection{Mean-squared errors backpropagation of weights in the path-activity framework}
The dynamics of a neural network during training are defined by the backpropagation algorithm, and we therefore naturally take the gradient descent update rule
\begin{equation}\label{eq:update_rule}
    \w{ij}{k}\longleftarrow \w{ij}{k} - \eta\pdv{\mathcal{L}}{\w{ij}{k}}
\end{equation}
as the defining equation of weight dynamics, and as the starting point for studying structure formation in deep neural networks. The first step in this direction is choosing a loss function $\mathcal{L}$, and here we specialise to the Mean Square Error (MSE) loss, given by
\begin{equation}
    \mathcal{L}_{\mathrm{MSE}}\left(Y^{(m)},\hat{Y}^{(m)}\right)=\frac{1}{2M}\dsum_{m=1}^{M}{\left(Y^{(m)}-\hat{Y}^{(m)}\right)^2}\mpt
\end{equation}
Note that we added the prefactor of $\frac{1}{2}$ which does not lead to any qualitative changes in the function, but is included here to aid simplifications of the equations that follow\footnote{Specifically, this factor will cancel when we compute the gradient of the loss function.} .
We start by considering a specific weight $\w{ab}{p}$, write the loss in the path-activity formalism,
\begin{align}
    \mathcal{L} = \frac{1}{2N}\dsum_{m=1}^{M}{\left(Y^{(m)} - \dsum_{i_j\in G}{X_{i_j}^{(m)}A_{i_j}^{(m)}\prod\limits_{k=1}^{H}{\w{i_j}{k}}}\right)^2}\mcm
\end{align}
and compute the gradient,
\begin{align*}\label{eq:path_gradient_intermediate}
    \pdv{\mathcal{L}}{\w{ab}{p}} &= \frac{1}{2N}\pdv{\w{ab}{p}}\left[\dsum_{m=1}^{M}{\left(Y^{(m)} - \dsum_{i_j\in G}{X_{i_j}^{(m)}A_{i_j}^{(m)}\prod\limits_{k=1}^{H}{\w{i_j}{k}}}\right)^2}\right] \\
    &= -\frac{1}{N}\dsum_{m=1}^{M}{\left(Y^{(m)}-\hat{Y}^{(m)}\right)} \left(\dsum_{i_j\in G}{X_{i_j}^{(m)}\pdv{\w{ab}{p}}\left[A_{i_j}^{(m)}\prod_{k=1}^{H}{\w{i_j}{k}}\right]}\right) \\
    &= -\frac{1}{N}\dsum_{m=1}^{M}{\left(Y^{(m)}-\hat{Y}^{(m)}\right)} \\ 
    &\qquad\quad\; \times \left(\dsum_{i_j\in G}{X_{i_j}^{(m)}\set{A_{i_j}^{(m)}\pdv{\w{ab}{p}}\prod_{k=1}^{H}{\w{i_j}{k}} + \pdv{A_{i_j}^{(m)}}{\w{ab}{p}}\prod_{k=1}^{H}{\w{i_j}{k}}}}\right)\mpt \numberthis
\end{align*}
The first derivative in the braces we compute as
\begin{align*}
    \pdv{\w{ab}{p}}\prod_{k=1}^{H}{\w{i_j}{k}} &= \dsum_{k=1}^{H}{\left(\left(\pdv{\w{ab}{p}}\w{i_j}{k}\right)\prod_{\substack{s=1 \\ s\neq p}}^{H}{\w{i_j}{(s)}}\right)} \\
    &= \dsum_{k=1}^{H}{\left(\deltafc{\w{i_j}{k} - \w{ab}{p}}\prod_{\substack{s=1 \\ s\neq p}}^{H}{\w{i_j}{(s)}}\right)}\mcm \numberthis
\end{align*}
and since there can be only one $k$, namely $k=p$, for which weight $\w{i_j}{k}$ can equal $\w{ab}{p}$, the summation over $k$ is redundant and we can write, after relabeling $s\to k$,
\begin{equation}
    \pdv{\w{ab}{p}}\prod_{k=1}^{H}{\w{i_j}{k}} = \prod_{\substack{k=1 \\ k\neq p}}^{H}{\w{i_j}{k}\deltafc{\w{i_j}{p} - \w{ab}{p}}}\mpt
\end{equation}
This $\delta$-function ensures that we sum over all paths using the unique weight connection $\w{ab}{p}$, since the path weights denoted by $\w{i_j}{k}$ are not unique.

To proceed from here, we need to consider the second term in the brackets, the gradient of the activity. For this we use the activity as defined in Eq.~\eqref{eq:path_activity}, such that
\begin{equation}
    \pdv{A_{i_j}^{(m)}}{\w{ab}{p}} =
    \pdv{\w{ab}{p}}\prod_{l=1}^{H}{\thetafc{\preact{i_j}{l}}}\mpt
\end{equation}
Now we note that since we derive with respect to a weight connecting layers $p-1$ and $p$, the pre-activations of the first $p-1$ terms of this product do not depend on that weight and can be considered as constants that can be pulled out of the derivative,
\begin{align*}
    \pdv{A_{i_j}^{(m)}}{\w{ab}{p}} &= \prod_{l=1}^{p-1}{\thetafc{\preact{i_j}{l}}}\pdv{\w{ab}{p}}\set{\prod_{l=p}^{H}{\thetafc{\preact{i_j}{l}}}} \\
    &= 
    \begin{aligned}[t]
        \prod_{l=1}^{p-1}{\thetafc{\preact{i_j}{l}}}\dsum_{s=p}^{H}\Bigg(\pdv{\w{ab}{p}} &\set{\thetafc{\preact{i_j}{s}}} \\ &\times \prod_{\substack{l=p \\ l\neq s}}^{H}{\thetafc{\preact{i_j}{l}}}\Bigg)
    \end{aligned} \\
    &= \prod_{l=1}^{p-1}{\thetafc{\preact{i_j}{l}}}\dsum_{s=p}^{H}\Bigg(\deltafc{\preact{i_j}{s}} \\ & \qquad\qquad\qquad\qquad\quad \times \pdv{\preact{i_j}{s}}{\w{ab}{p}} \prod_{\substack{l=p \\ l\neq s}}^{H}{\thetafc{\preact{i_j}{l}}}\Bigg)\mpt \numberthis
\end{align*}
In this equation the $\deltafc{\preact{i_j}{s}}$ will only be equal to $1$ if the pre-activation value is precisely equal to zero. The probability of this occurring with the floating point arithmetic used in numerical machine learning models is practically zero, which means that we can safely set
\begin{equation}
    \deltafc{\preact{i_j}{s}}=0\; \forall\, s\mpt
\end{equation}
The whole term then vanishes and we are left with
\begin{equation} \label{eq:path_activity_gradient}
    \pdv{\mathcal{L}}{\w{ab}{p}} = 
    -\frac{1}{N}\dsum_{m=1}^{M}{\left(Y^{(m)}-\hat{Y}^{(m)}\right)} \left(\dsum_{i_j\in G}\deltafc{\w{i_j}{p} - \w{ab}{p}} X_{i_j}^{(m)}A_{i_j}^{(m)}\prod_{\substack{k=1 \\ k\neq p}}^{H}{\w{i_j}{k}}\right)\mpt
\end{equation}
Note how this equation is structurally very similar to Eq.~\eqref{eq:path_activity_output}. The additional prefactor $\left(Y^{(m)}-\hat{Y}^{(m)}\right)$ tracks the training error, and within the summation over all paths a delta function has appeared, selecting only those paths that run through the weight that is being updated. Also, that weight is now excluded from the product of all weights along each path. As a sanity check, we recall the brief discussion of the dying ReLU problem, stating that when a node has a negative input for any sample of the data, it dies and its connected weights will no longer be updated. This exact phenomenon is also evident from Eq.~\eqref{eq:path_activity_gradient}, because whenever a weight connects to an inactive node, any path through that weight must be inactive. When that holds across all instances,
\begin{equation}
    A_{i_j}^{(m)}=0 \; \forall\, m\mcm
\end{equation}
we find that the gradient will equal zero, and the consequent reduction of Eq.~\eqref{eq:update_rule} to $\w{ij}{k}\longleftarrow \w{ij}{k}$ implies that weights remain constant.

\begin{figure}[!htb]
\centering
\includegraphics[scale=1.0]{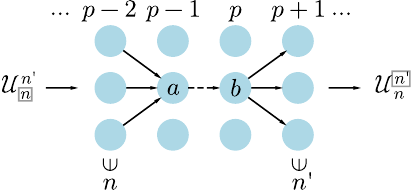}
\caption{The weight update of $\w{ab}{p}$ consists of all paths through this weight. The possible connections from layer $p-2$ to $p-1$ and from $p$ to $p+1$ are therefore restricted to those into and out of nodes $a$ and $b$ as in this exemplary 3-node-network. We specify these restricted connections by the nodes $n$ and $n'$ connecting to $a$ and $b$. Outside of this subset of layers, we do not have to impose any restrictions on the paths. We capture this part of the weight update by $\U{n}{n'}$, all paths into node $n^{(p-2)}$ and out of node $n'^{(p+1)}$.}
\label{fig:framework_schematic}
\end{figure}

We will use this equation as the starting point of further analytical calculations, so let us introduce some shorthand notation. In particular we want to give the weights left and right of the updating weight a more prominent place and pull them out of the product, the exact motivation of which will become clear later. To this end we split the summation over all paths in two. The first sum will include all paths into a node $n$ in layer $p-2$ and out of a node $n'$ in layer $p+1$, which we denote $G_{n}^{n'}(p)$. This set contains
\begin{equation}
    \left(\prod\limits_{i=0}^{p-3}{n_i}\right)\cdot\left(\prod\limits_{i=p+2}^{H}{n_i}\right) = \frac{n_0\Gamma}{n_{p-2}n_{p-1}n_{p}n_{p+1}}
\end{equation}
elements. The argument $p$ that indicates the layers in which $n$ and $n'$ are located is usually clear from the rest of the equation and we will therefore not explicitly state it and simply write $G_{n}^{n'}$. The second summation now has to run over all possible combinations $\set{n,n'}$, with
\begin{align}
    n &\in \set{1,\ldots,n_{p-2}}\mcm \\
    n' &\in \set{1,\ldots,n_{p+1}}\mcm
\end{align}
such that
\begin{equation}
    \dsum_{i_j\in G}\longrightarrow \dsum_{\set{n,n'}}\dsum_{i_j\in G_{n}^{n'}}\mpt
\end{equation}
The total number of terms in this double summation is now
\begin{equation}
    \frac{n_0\Gamma}{n_{p-2}n_{p-1}n_{p}n_{p+1}}\cdot n_{p-2}\cdot n_{p+1} =  \frac{n_0\Gamma}{n_{p-1}n_{p}}\mcm
\end{equation}
which is exactly the number of summands we expect if we would have imposed the restriction of the delta function, which forces us to use a certain weight, to the sum over all paths in $G$. Therefore, we can now remove the delta function completely at the cost of pulling the weights left and right of the updating weight out of the product,
\begin{equation}
    \deltafc{\w{i_j}{p} - \w{ab}{p}}\prod_{\substack{k=1 \\ k\neq p}}^{H}{\w{i_j}{k}} \longrightarrow \w{na}{p-1}\w{bn'}{p+1}\prod_{\substack{k=1 \\ k\neq p-1,p,p+1}}^{H}{\w{i_j}{k}}\mpt
\end{equation}
Altogether, the right hand side of Eq.~\eqref{eq:path_activity_gradient} then becomes
\begin{equation}
    -\frac{1}{N}\dsum_{m=1}^{M}{\left(Y^{(m)}-\hat{Y}^{(m)}\right)} \left(\dsum_{\set{n,n'}}\dsum_{i_j\in G_{n}^{n'}} X_{i_j}^{(m)}A_{i_j}^{(m)}\w{na}{p-1}\w{bn'}{p+1}\prod_{\substack{k=1 \\ k\neq p-1,p,p+1}}^{H}{\w{i_j}{k}}\right)\mpt
\end{equation}
To simplify further, we incorporate the summation restrictions imposed by the activity $A_{i_j}^{(m)}$ into the summation, and define $\mathcal{A}_{n}^{n'}(m)$ as a subset of $G_{n}^{n'}$ containing only the active paths in that set. This depends on the sample $m$. From now on we will drop this dependence fro the brevity of notation, and we have
\begin{equation}
    \dsum_{i_j\in G_{n}^{n'}}A_{i_j}^{(m)}\longrightarrow\dsum_{i_j\in \mathcal{A}_{n}^{n'}}\mpt
\end{equation}
With this we define
\begin{equation}\label{eq:U_definition}
    \U{n}{n'}\equiv \dsum_{i_j\in\mathcal{A}_{n}^{n'}}X_{i_j}^{(m')}\prod\limits_{\substack{k=1 \\ k\neq p-1,p, p+1}}^{H}{\w{i_j}{k}}
\end{equation}
and obtain the simplified form
\begin{equation}
    \pdv{\mathcal{L}}{\w{ab}{p}} = 
    -\frac{1}{N}\dsum_{m=1}^{M}{\left(Y^{(m)}-\hat{Y}^{(m)}\right)} \left(\dsum_{\set{n,n'}}\U{n}{n'}\w{na}{p-1}\w{bn'}{p+1}\right)\mpt
\end{equation}
For a structural understanding of this equation, we remind ourselves that a weight update in the path-activity formalism involves summing over all paths that run through the weight under consideration. Consider an exemplary network with three nodes in each layer, of which the part around $\w{ab}{p}$ is shown in Figure~\ref{fig:framework_schematic}. Instead of forcing paths through this weight by means of a delta function as we did in Eq.~\eqref{eq:path_activity_gradient}, we fixed the nodes $a$ and $b$ of its nearest neighbour weights respectively and pulled these weights out of the weight product. The nodes that specify these neighbouring weights we kept as summation parameters $n$ and $n'$. For this to work, we introduced the term $\U{n}{n'}$ which captures the contribution of the weight update into node $n^{(p-2)}$ and out of node $n^{(p+1)}$. By summing over all possible combinations $\set{n,n'}$ we again capture all paths through nodes $a$ and $b$ and obtained an equivalent description of the weight update.
The motivation behind this structure, is the observation that for two arbitrarily chosen weights, part of the respective increments will in general be the same. To understand this, let us consider the example of two weights that are in the same layer, $\w{ab}{p}$ and $\w{cd}{p}$. Taking a path through either of these weights means that we have to restrict to weight connections into and out of the nodes $a$ and $b$ or $c$ and $d$ respectively. If $a\neq c$ and $b\neq d$ then these nearest neighbour connections must be different. However, apart from this restriction all other weights along the respective paths can in principle be the same. This notation therefore simplifies the comparison of different weight updates, as it singles out their unique components, namely the nearest neighbour weights. When comparing weight updates of weights in different layers it is no longer just the nearest neighbour weights that are different. Nevertheless, with a slight modification we can still use the same notation to perform a quantitative analysis. 

We now denote by
\begin{align}
    \Delta^{(M)}\w{ab}{p} &\equiv \w{ab}{p}(\tau+1)-\w{ab}{p}(\tau) = -\eta\pdv{\mathcal{L}}{\w{ab}{p}} \\
    &\equiv \frac{\eta}{N}\dsum_{m=1}^{M}{\yerr} \left(\dsum_{\set{n,n'}}\U{n}{n'}\w{na}{p-1}\w{bn'}{p+1}\right)
\end{align}
the weight increment after one epoch $\tau$ of full batch gradient descent, with $\yerr\equiv Y^{(m)}-\hat{Y}^{(m)}$ the prediction error for sample $m$. Similarly,
\begin{align}
    \Delta^{(1)}\w{ab}{p} &\equiv \w{ab}{p}(\tau_{m+1})-\w{ab}{p}(\tau_m) \label{eq:S_coupled_weight_updates} \\
    &\equiv \eta\yerr \left(\dsum_{\set{n,n'}}\U{n}{n'}\w{na}{p-1}\w{bn'}{p+1}\right)
\end{align}
is the weight change in the case of full stochastic gradient descent, where weights are updated for each individual sample.

With this we are in a position to start studying how weights are interacting microscopically. Recall the observation that we can understand a weight update as a sum over all paths through this particular weight. This implies that when comparing updates for different weights, we can distinguish two cases. First there are all pairs of weights that are in the same layer or in successive layers but not connected to each other. Since it is impossible to take a path that runs through two weights in the same layer, or through unconnected weights in successive layers, the dynamics of these weights are in principle fully decoupled. Secondly, there are all pairs of connected weights in successive layers, and all those that are separated by one or more layers. For these pairs, it is possible for a path to run through both weights, and the corresponding weight dynamics are coupled. We will now first study the decoupled situation, and then introduce what happens in the case of having coupled weights in different layers. From the microscopic equations that govern these interactions we will infer the macroscopic consequences, and show how large scale structures arise.

\subsection{Extension to other activation functions}\label{sec:S_approximating_act_fct}
\begin{figure}[!htb]
\centering
\includegraphics[scale=1.0]{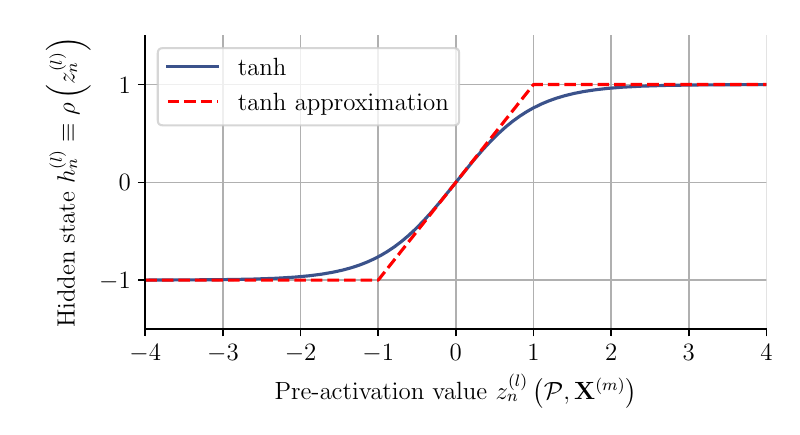}
\caption{Piecewise linear approximation of the hyperbolic tangent. We can approximate the hyperbolic tangent using the three domains $(-\infty,-1)$, $[-1,1]$, and $(1,\infty)$, where we set $\tanh(x)=\set{-1,x,1}$ respectively.}
\label{fig:tanh_approx}
\end{figure}

Let $\actfc{}$ be an arbitrary activation function. We know that the reason $\mathtext{ReLU}$ suits itself for the path-activity formalism, is the fact that it is piecewise linear. This allows us to represent its activation as the constant slopes of 0 and 1 of the two linear parts of the function. One idea could therefore be to approximate a general activation function $\actfc{}$ by a piecewise linear function,
\begin{equation}
    \actfc{}(x)\approx \dsum_{i=1}^{n}{\left(\beta_i x + \zeta_i\right)\mathbf{1}_{L_i}(x)} \mpt
\end{equation}
Here $n$ is the number of linear pieces by which we approximate the function, and $\mathbf{1}_{L_i}$ is the indicator function on the interval $L_i$ on which the function has the linear form $\beta_i x + \zeta_i$, for slope $\beta_i$ and offset $\zeta_i$. In the limit of infinitely small intervals we recover the original, fully continuous function. An example of this approximation for the hyperbolic tangent activation with three intervals is shown in Figure~\ref{fig:tanh_approx}.
The activation state of a node is now no longer captured by a single multiplicative step function, but two separate parts. First a multiplicative piecewise constant function representing a set of discrete values by which the input is multiplied,
\begin{equation}
   A_{\mathtext{mult}}(x) = \dsum_{i=1}^{n}{\beta_i\mathbf{1}_{L_i}(x)} \mpt
\end{equation}
Second, an additive part that acts as a bias to the multiplicative activation,
\begin{equation}
    A_{\mathtext{bias}}(x) =  \dsum_{i=1}^{n}{\zeta_i\mathbf{1}_{L_i}(x)} \mpt
\end{equation}
As an example, we can write
\begin{equation}
    \mathtext{ReLU}(x) = \left(0\cdot x + 0\right)\mathbf{1}_{(-\infty,0)}(x) + \left(1\cdot x + 0\right)\mathbf{1}_{[0,\infty)} \mcm
\end{equation}
and
\begin{equation}
    \pdv{\mathtext{ReLU}(x)}{x} = 0\cdot\mathbf{1}_{(-\infty,0)}(x) + 1\cdot \mathbf{1}_{[0,\infty)} \mpt
\end{equation}
We now see that it is also the vanishing bias terms that make $\mathtext{ReLU}$ particularly well-suited for the activation description. In general, instead of having a binary activation state, we would thus have a discrete set of numbers that characterise the activity of a node. We can increase the precision of the approximation by varying the number $n$ of different activation levels we consider. With this intermediate step, we believe it should be possible to extend the framework to different activation functions. 

\section{Feedback loops between weights in different layers}\label{sec:Feedback_loops}
Now that we understand the microscopic dynamics of weights in the same layer, the next  step is to study the dynamics of weights in different layers, for which the updates are no longer independent, but coupled to each other.

\subsection{Microscopic interlayer weight dynamics}\label{sec:FL_micr}
Microscopically, we want to understand the coupled updating of two arbitrary weights, $\w{ab}{p}$ and $\w{cd}{k}$, $p\neq k$, for which a certain subset of paths is identical. A full treatment of this scenario requires separating three cases: connected weights in successive layers, weights separated by one layer, and weights separated by two or more layers.

\subsubsection{Coupling between connected weights in successive layers}
\begin{figure}[!htb]
\centering
    \sidesubfloat[]{\includegraphics[scale=1.0]{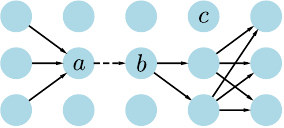}\label{subfig:conn_no_coupling}} \quad%
    \sidesubfloat[]{\includegraphics[scale=1.0]{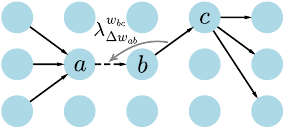}\label{subfig:conn_coupling}}
    \caption{For connected weights $\w{ab}{p}$ and $\w{bc}{p+1}$ in successive layers, the weight update of the former consists of two parts. \protect\subref{subfig:conn_no_coupling} First we have all paths that \textbf{do not} run through the connected weight $\w{bc}{p+1}$, corresponding to the first term in the brackets of Eq.~\eqref{eq:coupling_bc_to_ab}. \protect\subref{subfig:conn_coupling} Secondly, we have all paths that \textbf{do} use the connection $\w{bc}{p+1}$, leading to a feedback from that weight to the updated weight $\w{ab}{p}$, generated by the second term in the brackets of Eq.~\eqref{eq:coupling_bc_to_ab}. This feedback is quantified by a coupling $\lambda_{\Delta\w{ab}{p}}^{\w{bc}{p+1}}$.}\label{fig:connected_weights}
\end{figure}

Let $\w{ab}{p}$ and $\w{bc}{p+1}$ be two weights that both connect to node $b^{(p)}$. We start by rewriting their respective weight updates,
\begin{align*}
    \Delta^{(1)}\w{ab}{p} &= \eta\yerr \left(\dsum_{\set{n,n'}}\U{n}{n'}\w{na}{p-1}\w{bn'}{p+1}\right) \\
    &= \eta\yerr\dsum_{n}\left[\dsum_{n'\neq c}\U{n}{n'}\w{na}{p-1}\w{bn'}{p+1} + \U{n}{c}\w{na}{p-1}\w{bc}{p+1}\right] \mcm \numberthis\label{eq:coupling_bc_to_ab} \\
    \Delta^{(1)}\w{bc}{p+1} &= \eta\yerr \left(\dsum_{\set{n,n'}}\U{n}{n'}\w{nb}{p}\w{cn'}{p+2}\right) \\
    &= \eta\yerr\dsum_{n'}\left[\dsum_{n\neq a}\U{n}{n'}\w{nb}{p}\w{cn'}{p+2} + \U{a}{n'}\w{ab}{p}\w{cn'}{p+2}\right] \numberthis \mpt
\end{align*}
In these equations we excluded $\w{ab}{p}$ and $\w{bc}{p+1}$ respectively from the summation and added these terms individually. This clarifies the precise role each of these weights plays in the updating of the other. We see that the weight updates are coupled, as $\Delta^{(1)}\w{ab}{p}$ depends on the value of $\w{bc}{p+1}$ and vice versa. The coupling constants $\lambda_{\mu}^{\nu}$ that govern the effect of some quantity $\nu$ on another quantity $\mu$ are 
\begin{align}
    \lambda_{\Delta\w{ab}{p}}^{\w{bc}{p+1}} &= \eta\yerr\dsum_{n}{\U{n}{c}\w{na}{p-1}} \mcm \\
    \lambda_{\Delta\w{bc}{p+1}}^{\w{ab}{p}} &= \eta\yerr\dsum_{n'}{\U{a}{n'}\w{cn'}{p+2}} \mpt
\end{align}
Figure~\ref{fig:connected_weights} gives a visual interpretation of the two terms in these update equations for our toy network with a width of 3 nodes, taking Eq.~\eqref{eq:coupling_bc_to_ab} as an example. We can understand the first term as a sum over all paths through $\w{ab}{p}$ that do not use the connection $\w{bc}{p+1}$, as shown in Figure~\ref{subfig:conn_no_coupling}. In this case, there is no coupling between these weights. The second term, represented by Figure~\ref{subfig:conn_coupling}, singles out all paths that run through both weights, resulting in a dependence of the update on $\w{bc}{p+1}$, characterised by a coupling $\lambda$. Together, we thus have pairs of coupled update equations for all connected weights in successive layers, such that each weight feeds back on the update of the other weight.
\floatsetup[figure]{style=plain,subcapbesideposition=top}

\subsubsection{Coupling between weights separated by one or more layers}
\begin{figure}[!htb]
    \sidesubfloat[]{\includegraphics[scale=1.0]{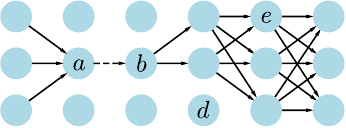}\label{subfig:sep_no_coupling_1}} \\[\baselineskip]%
    \sidesubfloat[]{\includegraphics[scale=1.0]{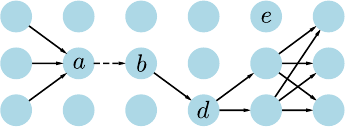}\label{subfig:sep_no_coupling_2}} \\[\baselineskip]%
    \sidesubfloat[]{\includegraphics[scale=1.0]{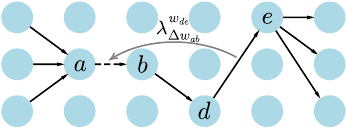}\label{subfig:sep_coupling}}
    \caption{Separated weight coupling. Weights $\w{ab}{p}$ and $\w{de}{p+2}$ separated by one layer are always coupled, as the weight update splits in three terms. \protect\subref{subfig:sep_no_coupling_1} This figure represents the first term in the brackets of Eq.~\eqref{eq:separated_coupling_de_to_ab}, for any of the paths that do not run through node $d^{(p+1)}$, it is impossible to use weight connection $\w{de}{p+2}$. \protect\subref{subfig:sep_no_coupling_2} When we do use that node, it is still possible to continue from there via paths that do not incorporate node $e^{(p+2)}$, amounting to the second term in Eq.~\eqref{eq:separated_coupling_de_to_ab}, again without coupling. \protect\subref{subfig:sep_coupling} The third term in the brackets captures the coupling $\lambda_{\Delta\w{ab}{p}}^{\w{de}{p+2}}$ by considering all paths through both weights.}\label{fig:separated_weights}
\end{figure}

In this case we look at the weights $\w{ab}{p}$ and $\w{de}{p+2}$, with none of the nodes overlapping. Again we rewrite the increment,
\begin{align*}\label{eq:separated_coupling_de_to_ab}
    \Delta^{(1)}\w{ab}{p} &= \eta\yerr \left(\dsum_{\set{n,n'}}\U{n}{n'}\w{na}{p-1}\w{bn'}{p+1}\right) \\
    &= \eta\yerr\dsum_{n}\left[\dsum_{n'\neq d}\U{n}{n'}\w{na}{p-1}\w{bn'}{p+1} + \U{n}{d}\w{na}{p-1}\w{bd}{p+1}\right] \\
    &= 
    \begin{aligned}[t]
        \eta\yerr\dsum_{n}&\Bigg[\dsum_{n'\neq d}\U{n}{n'}\w{na}{p-1}\w{bn'}{p+1} \\ & + \left(\dsum_{n"}\U{n}{n"}\w{dn"}{p+2}\right) \w{na}{p-1}\w{bd}{p+1}\Bigg]
    \end{aligned}
    \\
    &=
    \begin{aligned}[t]
        \eta\yerr\dsum_{n}&\Bigg[\dsum_{n'\neq d}\U{n}{n'}\w{na}{p-1}\w{bn'}{p+1} + \\ &\left(\dsum_{n"\neq e}\U{n}{n"}\w{dn"}{p+2} + \U{n}{e}\w{de}{p+2}\right) \w{na}{p-1}\w{bd}{p+1}\Bigg]
    \end{aligned}
    \\
    &= \eta\yerr\dsum_{n}\Bigg[\dsum_{n'\neq d}\U{n}{n'}\w{na}{p-1}\w{bn'}{p+1} \\ &\quad+\w{na}{p-1}\w{bd}{p+1}\dsum_{n"\neq e}\U{n}{n"}\w{dn"}{p+2} + \U{n}{e}\w{na}{p-1}\w{bd}{p+1}\w{de}{p+2} \Bigg]\mpt \numberthis
\end{align*}
For $\w{de}{p+2}$ we derive analogously
\begin{align*}\label{eq:separated_coupling_ab_to_de}
    \Delta^{(1)}\w{de}{p+2} &= \eta\yerr \left(\dsum_{\set{n,n'}}\U{n}{n'}\w{nd}{p+1}\w{en'}{p+3}\right) \\
    &= \eta\yerr\dsum_{n'}\Bigg[\dsum_{n\neq b}\U{n}{n'}\w{nd}{p+1}\w{en'}{p+3} \\ &\quad+\w{bd}{p+1}\w{en'}{p+3}\dsum_{n"\neq a}\U{n"}{n'}\w{n"b}{p+2} + \U{a}{n'}\w{ab}{p}\w{bd}{p+1}\w{en'}{p+3} \Bigg]\mpt \numberthis
\end{align*}
The coupling constants are now given by
\begin{align}
    \lambda_{\Delta\w{ab}{p}}^{\w{de}{p+2}} &= \eta\yerr\dsum_{n}{\U{n}{e}\w{na}{p-1}\w{bd}{p+1}} \mcm \\
    \lambda_{\Delta\w{de}{p+2}}^{\w{ab}{p}} &= \eta\yerr\dsum_{n'}{\U{a}{n'}\w{bd}{p+1}\w{en'}{p+3}} \mpt
\end{align}
The update equations Eq.~\eqref{eq:separated_coupling_de_to_ab} and ~\eqref{eq:separated_coupling_ab_to_de} now contain three terms, corresponding to the three sketches in Figure~\ref{fig:separated_weights}. These terms arise from the fact that in order to reach the connection belonging to $\w{de}{p+2}$, we must continue from node $b^{(p)}$ to node $d^{(p+1)}$, enforcing a first division of the full summation over all paths in the second line of these equations. Secondly, even when we are in node $d^{(p+1)}$, we can still decide to continue our journey to any other node than $e^{(p+2)}$, hence creating a second split in the summation, the computational steps of which are given in lines 3 and 4 of the equations, before arriving at the final expression with three terms. The first term thus accounts for all paths through nodes $a^{(p-1)}$ and $b^{(p)}$ but not $d^{(p+1)}$, as indicated in Figure~\ref{subfig:sep_no_coupling_1}. Figure~\ref{subfig:sep_no_coupling_2} entails all paths that actually use $\w{bd}{p+1}$, but fail to go through node $e^{(p+2)}$. The third and final term, Figure~\ref{subfig:sep_coupling}, gives us the desired coupling by successfully using nodes $b^{(p)}$, $d^{(p+1)}$, and $e^{(p+2)}$. We now want to understand how strong the couplings are relative to each other for the two scenarios of directly connected weights in successive layers, and weights separated by one layer. To this end, we compute the ratio,
\floatsetup[figure]{style=plain,subcapbesideposition=top}
\begin{align}
    \frac{\lambda_{\Delta\w{ab}{p}}^{\w{de}{p+2}}}{\lambda_{\Delta\w{ab}{p}}^{\w{bc}{p+1}}} &= \frac{\eta\yerr\dsum_{n}{\U{n}{e}\w{na}{p-1}\w{bd}{p+1}}}{\eta\yerr\dsum_{n}{\U{n}{c}\w{na}{p-1}}} \\
    &= \frac{\dsum_{n}{\U{n}{e}\w{na}{p-1}\w{bd}{p+1}}}{\dsum_{n}{\U{n}{c}\w{na}{p-1}}} \\
    &= \frac{\w{bd}{p+1}\dsum_{n}{\U{n}{e}\w{na}{p-1}}}{\dsum_{n}{\left(\dsum_{n"\neq e}{\U{n}{n"}\w{cn"}{p+2}}+\U{n}{e}\w{ce}{p+2}\right)\w{na}{p-1}}}\mpt
\end{align}
To proceed from here, we assume that
\begin{align}
    \U{n}{n"} &\approx \U{n}{e} \mcm \\
    \w{cn"}{p+2} &\approx \w{ce}{p+2} \mcm
\end{align}
such that they are independent of the summation parameter $n"$, reducing the sum to a multiplication with a factor of $n_{p+2}-1$. We can then group the two terms in the brackets of the denominator together,
\begin{align}
    \frac{\lambda_{\Delta\w{ab}{p}}^{\w{de}{p+2}}}{\lambda_{\Delta\w{ab}{p}}^{\w{bc}{p+1}}} &\approx \frac{\w{bd}{p+1}\dsum_{n}{\U{n}{e}\w{na}{p-1}}}{\dsum_{n}{\left(\left(n_{p+2}-1\right){\U{n}{e}\w{ce}{p+2}}+\U{n}{e}\w{ce}{p+2}\right)\w{na}{p-1}}} \\
    &= \frac{\w{bd}{p+1}\dsum_{n}{\U{n}{e}\w{na}{p-1}}}{n_{p+2}\w{ce}{p+2}\dsum_{n}{\U{n}{e}\w{na}{p-1}}} \\
    &= \frac{1}{n_{p+2}}\frac{\w{bd}{p+1}}{\w{ce}{p+2}} \mpt
\end{align}
If we set $\w{bd}{p+1}=\w{ce}{p+2}$, we thus find that the successive layer coupling is stronger by factor $n_{p+2}$. We can understand this scaling in a heuristic manner, by considering how the number of paths reduces when fixing a certain number of nodes. For the nearest neighbour coupling, we had to fix three nodes in consecutive layers. Following our calculations above, let these nodes be in layers $p-1$, $p$, and $p+1$, then the number of paths we can take through them is
\begin{equation}
    \frac{n_0\Gamma}{n_{p-1}n_{p}n_{p+1}}\mpt
\end{equation}
In the separated case, we had to fix four nodes, one in each of the same layers as above, and an additional one in layer $p+2$, which reduces the number of paths to
\begin{equation}
    \frac{n_0\Gamma}{n_{p-1}n_{p}n_{p+1}n_{p+2}}\mpt
\end{equation}
By comparison we see that the ratio of these numbers is precisely the scaling of $\frac{1}{n_{p+2}}$. As the number of possible paths, which defines the maximal magnitude of a weight update, reduces, so does the maximal influence of two weights on each others update.


With this heuristic argument, it is now easy to understand that the coupling of weights separated by two or more layers will have the exact same strength as for a one layer separation. Namely, no matter how far apart, we always have to fix a constant number of precisely four nodes to take a path through both. The coupling of an arbitrary weight $\w{fg}{k}$ to $\w{ab}{p}$, $k-p\geq 2$ as compared to the nearest neighbour coupling with a weight $\w{bc}{p+1}$ will thus scale as
\begin{equation}
    \frac{\lambda_{\Delta\w{ab}{p}}^{\w{fg}{k}}}{\lambda_{\Delta\w{ab}{p}}^{\w{bc}{p+1}}} \sim \frac{1}{n_{k}} \mpt
\end{equation}
We conclude that the connected nearest neighbour coupling is unique and strong, whereas any two other weights in the network separated by at least one layer experience a coupling that is always of the same order. Nearest neighbour effects should thus dominate the dynamics of the network.

To leading order, we thus find that there is feedback between connected weights in successive layers, such that a large weight will make all its neighbours large, and as the neighbours get larger, this feeds back to the first weight as well. The larger both connected weights are, the stronger they affect each others weight update. 

\section{Intralayer dynamics of nodal connectivity}
\subsection{Definition of nodal connectivity}
We now use the formalism introduced above to derive the dynamics of the connectivities, which are defined as the product of in- and outgoing weight fractions,
\begin{align}
    r_{\text{abs}}\left(n_j,l\right) &\equiv \Omega_{\text{in}}\left(n_j,l\right) \cdot \Omega_{\text{out}}\left(n_j,l\right) \\
    &=\frac{\sum_i{|w_{ij}^{(l)}|}}{\sum_{i,j}{|w_{ij}^{(l)}|}} \cdot \frac{\sum_k{|w_{jk}^{(l+1)}|}}{\sum_{j,k}{|w_{jk}^{(l+1)}|}}\mpt
\end{align}
The reason for using absolute values of weights, is that a priori we cannot distinguish between the importance of a negative weight as compared to a positive weight for the local functioning of the network. We therefore take absolute values to give all weights equal sign and not let this sign affect the local morphology of the network.

\subsection{Dynamics of nodal connectivity without explicit adjacent layer coupling}
Using the chain rule we now derive the continuous time evolution of the connectivity. In reality we of course have discrete weight and thus connectivity updates, this is therefore an approximation that only holds for small enough weight update in each discrete step.
\begin{equation}
    \dv{t} r_{\text{abs}}\left(n_j,l\right) = \Omega_{\text{in}}\left(n_j,l\right) \cdot \dv{t} \Omega_{\text{out}}\left(n_j,l\right) + \dv{t} \Omega_{\text{in}}\left(n_j,l\right) \cdot \Omega_{\text{out}}\left(n_j,l\right)\mpt
    \label{eq:r_dynamics}
\end{equation}
For the time derivatives of the weight fractions we find
\begin{align}
    &\dv{t} \Omega_{\text{in}}\left(n_j,l\right) = \frac{\sum_i{\dv{t}|w_{ij}^{(l)}|}}{\sum_{i,j}{|w_{ij}^{(l)}|}} - \frac{\sum_i{|w_{ij}^{(l)}|}\cdot \sum_{i,j}{\dv{t}|w_{ij}^{(l)}|}}{\left(\sum_{i,j}{|w_{ij}^{(l)}|}\right)^2} \\
    &\equiv \frac{1}{W^{(l)}}\left(\sum_i{\dv{t}|w_{ij}^{(l)}|} - \Omega_{j}^{\text{in}}\sum_{i,j}{\dv{t}|w_{ij}^{(l)}|}\right) \\
    &\dv{t} \Omega_{\text{out}}\left(n_j,l\right) = \frac{\sum_k{\dv{t}|w_{jk}^{(l+1)}|}}{\sum_{j,k}{|w_{jk}^{(l+1)}|}} - \frac{\sum_k{|w_{jk}^{(l+1)}|}\cdot \sum_{j,k}{\dv{t}|w_{jk}^{(l+1)}|}}{\left(\sum_{j,k}{|w_{jk}^{(l+1)}|}\right)^2} \\
    &\equiv \frac{1}{W^{(l+1)}}\left(\sum_k{\dv{t}|w_{jk}^{(l+1)}|} - \Omega_{j}^{\text{out}}\sum_{j,k}{\dv{t}|w_{jk}^{(l+1)}|}\right)\mcm
\end{align}
where $W^{(l)}$ denotes the total amount of absolute weight connecting layers $l-1$ and $l$. Plugging this into Eq.~\eqref{eq:r_dynamics} for the connectivity dynamics we get
\begin{align}
    \dv{t} r_{\text{abs}}\left(n_j,l\right) &= \Omega_{j}^{\text{in}}\frac{1}{W^{(l+1)}}\left(\sum_k{\dv{t}|w_{jk}^{(l+1)}|} - \Omega_{j}^{\text{out}}\sum_{j,k}{\dv{t}|w_{jk}^{(l+1)}|}\right) \\&+ \Omega_{j}^{\text{out}}\frac{1}{W^{(l)}}\left(\sum_i{\dv{t}|w_{ij}^{(l)}|} - \Omega_{j}^{\text{in}}\sum_{i,j}{\dv{t}|w_{ij}^{(l)}|}\right) \\
    &= \frac{1}{W^{(l+1)}}\left(\Omega_{j}^{\text{in}}\sum_k{\dv{t}|w_{jk}^{(l+1)}|} - r_j\sum_{j,k}{\dv{t}|w_{jk}^{(l+1)}|}\right) \\&+ \frac{1}{W^{(l)}}\left(\Omega_{j}^{\text{out}}\sum_i{\dv{t}|w_{ij}^{(l)}|} - r_j\sum_{i,j}{\dv{t}|w_{ij}^{(l)}|}\right) \\
    &= \frac{1}{W^{(l+1)}}\left(\Omega_{j}^{\text{in}}\sum_k{\text{sgn}(w_{jk}^{(l+1)})\dv{t}w_{jk}^{(l+1)}} - r_j\sum_{j,k}{\text{sgn}(w_{jk}^{(l+1)})\dv{t}w_{jk}^{(l+1)}}\right) \\&+ \frac{1}{W^{(l)}}\left(\Omega_{j}^{\text{out}}\sum_i{\text{sgn}(w_{ij}^{(l)})\dv{t}w_{ij}^{(l)}} - r_j\sum_{i,j}{\text{sgn}(w_{ij}^{(l)})\dv{t}w_{ij}^{(l)}}\right)\mpt
\end{align}
From here we can continue, still exactly, by employing the expression we derived earlier for the discrete weight updates in the path-activity framework, Eq.~\eqref{eq:S_coupled_weight_updates},
\begin{align}
    &= \frac{1}{W^{(l+1)}}\left(\Omega_{j}^{\text{in}}\sum_k{\text{sgn}(w_{jk}^{(l+1)})\sum_{n,m}{\mathcal{U}_n^m w_{nj}^{(l)}w_{km}^{(l+2)}}} - r_j\sum_{j,k}{\text{sgn}(w_{jk}^{(l+1)})\sum_{n,m}{\mathcal{U}_n^m w_{nj}^{(l)}w_{km}^{(l+2)}}}\right) \\&+ \frac{1}{W^{(l)}}\left(\Omega_{j}^{\text{out}}\sum_i{\text{sgn}(w_{ij}^{(l)})\sum_{o,p}{\mathcal{U}_o^p w_{oi}^{(l-1)}w_{jp}^{(l+1)}}} - r_j\sum_{i,j}{\text{sgn}(w_{ij}^{(l)})\sum_{o,p}{\mathcal{U}_o^p w_{oi}^{(l-1)}w_{jp}^{(l+1)}}}\right)\mpt
\end{align}
In this step we approximated the continuous weight updates by their discrete counterparts from real neural networks. To simplify this expression we now realise that $\sum_{n}{w_{nj}^{(l)}}\sim\sum_{n}{\abs{w_{nj}^{(l)}}}=\Omega_{j}^{\text{in}}W^{(l)}$, and similar for the other sums. In other words, the sum over all signed weights must be smaller than or equal to the sum over all absolute weights, and we will define the proportionality factor down below. However, to use this relation, we first need to deal with the nodal dependency of the couplings $\mathcal{U}$. To that end, we note that initially the network is in a near-homogeneous state, in which case this term can be approximated to be independent of the nodes it connects, i.e. $n$ and $m$ or $o$ and $p$. We thus approximate close to the homogeneous state,
\begin{align}
    &\approx \frac{\mathcal{U}_R}{W^{(l+1)}}\left(\Omega_{j}^{\text{in}}\sum_k{\text{sgn}(w_{jk}^{(l+1)})\sum_{n}{w_{nj}^{(l)}}\sum_{m}{w_{km}^{(l+2)}}} - r_j\sum_{j,k}{\text{sgn}(w_{jk}^{(l+1)})\sum_{n}{w_{nj}^{(l)}}\sum_{m}{w_{km}^{(l+2)}}}\right) \\&+ \frac{\mathcal{U}_L}{W^{(l)}}\left(\Omega_{j}^{\text{out}}\sum_i{\text{sgn}(w_{ij}^{(l)})\sum_{o}{w_{oi}^{(l-1)}}\sum_{p}{w_{jp}^{(l+1)}}} - r_j\sum_{i,j}{\text{sgn}(w_{ij}^{(l)})\sum_{o}{w_{oi}^{(l-1)}}\sum_{p}{w_{jp}^{(l+1)}}}\right)\mpt
\end{align}
Now we use our insight from above to write $\sum_{n}{w_{nj}^{(l)}}\sim c_{j}^{\text{in}}\sum_{n}{\abs{w_{nj}^{(l)}}}=c_{j}^{\text{in}}\Omega_{j}^{\text{in}}W^{(l)}$ and similarly $\sum_{p}{w_{jp}^{(l+1)}}\sim c_{j}^{\text{out}}\sum_{p}{\abs{w_{jp}^{(l+1)}}}=c_{j}^{\text{out}}\Omega_{j}^{\text{out}}W^{(l+1)}$, which leads to
\begin{align}
    &\approx \frac{\mathcal{U}_R}{W^{(l+1)}}\left(\Omega_{j}^{\text{in}}\sum_k{\text{sgn}(w_{jk}^{(l+1)})c_j^{\text{in}}W^{(l)}\Omega_j^{\text{in}}\sum_{m}{w_{km}^{(l+2)}}} - r_j\sum_{j,k}{\text{sgn}(w_{jk}^{(l+1)})c_j^{\text{in}}W^{(l)}\Omega_j^{\text{in}}\sum_{m}{w_{km}^{(l+2)}}}\right) \\&+ \frac{\mathcal{U}_L}{W^{(l)}}\left(\Omega_{j}^{\text{out}}\sum_i{\text{sgn}(w_{ij}^{(l)})\sum_{o}{w_{oi}^{(l-1)}}c_j^{\text{out}}W^{(l+1)}\Omega_j^{\text{out}}} - r_j\sum_{i,j}{\text{sgn}(w_{ij}^{(l)})\sum_{o}{w_{oi}^{(l-1)}}c_j^{\text{out}}W^{(l+1)}\Omega_j^{\text{out}}}\right) \mpt
\end{align}
We can now pull the $j$-dependent terms out of the sums and use the relations $\Omega_j^{\text{out}} = \frac{r_j}{\Omega_j^{\text{in}}}$ and $\Omega_j^{\text{in}} = \frac{r_j}{\Omega_j^{\text{out}}}$. We also note that the second term with the summation over all $j$ can be split up into a term where $j$ equals the $j$ that we are considering, and all other nodes, and this gives us
\begin{align}
    &= \frac{\mathcal{U}_R}{W^{(l+1)}}\bigg(\frac{\Omega_{j}^{\text{in}}}{\Omega_{j}^{\text{out}}}r_j c_j^{\text{in}}W^{(l)}\sum_k{\text{sgn}(w_{jk}^{(l+1)})\sum_{m}{w_{km}^{(l+2)}}} - r_j \frac{r_j}{\Omega_j^{\text{out}}}c_j^{\text{in}}W^{(l)}\sum_{k}{\text{sgn}(w_{jk}^{(l+1)})\sum_{m}{w_{km}^{(l+2)}}} \\ &- r_j\sum_{m\neq j,k}{\text{sgn}(w_{jk}^{(l+1)})c_j^{\text{in}}W^{(l)}\Omega_j^{\text{in}}\sum_{m}{w_{km}^{(l+2)}}}\bigg) \\&+ \frac{\mathcal{U}_L}{W^{(l)}}\bigg(\frac{\Omega_{j}^{\text{out}}}{\Omega_j^{\text{in}}}r_jc_j^{\text{out}}W^{(l+1)}\sum_i{\text{sgn}(w_{ij}^{(l)})\sum_{o}{w_{oi}^{(l-1)}}} - r_j \frac{r_j}{\Omega_j^{\text{in}}}c_j^{\text{out}}W^{(l+1)}\sum_{i}{\text{sgn}(w_{ij}^{(l)})\sum_{o}{w_{oi}^{(l-1)}}} \\& - r_j\sum_{i,m\neq j}{\text{sgn}(w_{im}^{(l)})\sum_{o}{w_{oi}^{(l-1)}}c_m^{\text{out}}W^{(l+1)}\Omega_m^{\text{out}}}\bigg) \mpt
\end{align}
now we group the first two terms together,
\begin{align}
    &= \frac{\mathcal{U}_R}{W^{(l+1)}}\left(\left(\Omega_j^{\text{in}}r_j -r_j^2\right)\frac{1}{\Omega_j^{\text{out}}}c_j^{\text{in}}W^{(l)}\sum_k{\text{sgn}(w_{jk}^{(l+1)})\sum_{m}{w_{km}^{(l+2)}}} - r_j\sum_{m\neq j,k}{\text{sgn}(w_{mk}^{(l+1)})c_m^{\text{in}}W^{(l)}\Omega_m^{\text{in}}\sum_{m}{w_{km}^{(l+2)}}}\right) \\&+
    \frac{\mathcal{U}_L}{W^{(l)}}\left(\left(\Omega_j^{\text{out}}r_j -r_j^2\right)\frac{1}{\Omega_j^{\text{in}}}c_j^{\text{out}}W^{(l+1)}\sum_i{\text{sgn}(w_{ij}^{(l)})\sum_{o}{w_{oi}^{(l-1)}}} - r_j\sum_{i,m\neq j}{\text{sgn}(w_{im}^{(l)})\sum_{o}{w_{oi}^{(l-1)}}c_m^{\text{out}}W^{(l+1)}\Omega_m^{\text{out}}}\right) \mpt
    \label{eq:r_dynamics_before_approx}
\end{align}
Finally, ignoring the specific dependence on interlayer couplings for now, define
\begin{align}
    c_j^R &\equiv \frac{\mathcal{U}_R}{W^{(l+1)}}c_j^{\text{in}}W^{(l)}\sum_k{\text{sgn}(w_{jk}^{(l+1)})\sum_{m}{w_{km}^{(l+2)}}}\mcm \\
    c_j^L &\equiv \frac{\mathcal{U}_L}{W^{(l)}}c_j^{\text{out}}W^{(l+1)}\sum_i{\text{sgn}(w_{ij}^{(l)})\sum_{o}{w_{oi}^{(l-1)}}}\mpt
\end{align}
We now make a final approximation, namely that $\Omega_j^{\text{in/out}}\sim\sqrt{r_j}$. This follows from the observation that the ingoing and outgoing weight fractions are strongly correlated, and follow a near perfect linear slope, as shown in figure~\ref{fig:2}e. Therefore, $r_j=\Omega_j^{\text{in}}\cdot \Omega_j^{\text{out}}\approx \left(\Omega_j^{\text{in/out}}\right)^2$, from which the approximation directly follows. Plugging this into the expression above, we obtain the final result for the connectivity dynamics without explicit interlayer coupling,
\begin{align}
    \dv{r_j}{t}& \approx \left((r_j - r_j\sqrt{r_j})c_j^R - r_j\sum_{m\neq j}{\sqrt{r_m}c_m^R}\right) + \left((r_j - r_j\sqrt{r_j})c_j^L - r_j\sum_{m\neq j}{\sqrt{r_m}c_m^L}\right) \\
    &= r_j(1-\sqrt{r_j})c_j - r_j\sum_{m\neq j}{\sqrt{r_m}c_m}
    \label{eq:coarse_grained_r_dynamics}
\end{align}
where now $c_j\equiv c_j^R + c_j^L$. Notice how the left-right symmetry of the network is reflected in the two, symmetric terms of the first approximate equality. The $c_j$ are initially all roughly equal with small perturbations when the network is still in its homogeneous state. As these $c_j$ can be interpreted as growth rates, these small perturbations leads, on small time scales, to relative growing and shrinking of nodes. \\ \newline

\subsection{Dynamics of nodal connectivities with explicit adjacent layer coupling}
We now start from Eq.~\eqref{eq:r_dynamics_before_approx} and study to highest order the exact dependency of the dynamics on couplings to adjacent layers. To this end, let us define $q_{ij}\equiv\text{sgn}\left(w_{ij}\right)=\pm 1$ and we again realise that $\sum_{m}{w_{km}^{(l+2)}}\sim c_{k,l+1}^{\text{out}}\sum_{m}{\abs{w_{km}^{(l+2)}}}=c_{k,l+1}^{\text{out}}\Omega_{k,l+1}^{\text{out}}W^{(l+2)}$, and $\sum_{o}{w_{oi}^{(l-1)}}\sim c_{i,l-1}^{\text{in}}\sum_{o}{\abs{w_{oi}^{(l-1)}}}=c_{i,l-1}^{\text{in}}\Omega_{i,l-1}^{\text{in}}W^{(l-1)}$, leading to
\begin{align}
    &\approx \frac{\mathcal{U}_R}{W^{(l+1)}}\left(\left(\Omega_j^{\text{in}}r_j -r_j^2\right)\frac{1}{\Omega_j^{\text{out}}}c_j^{\text{in}}W^{(l)}\sum_k{q_{jk}c_{k,l+1}^{\text{out}}\Omega_{k,l+1}^{\text{out}}W^{(l+2)}} - r_j\sum_{m\neq j,k}{q_{mk}c_m^{\text{in}}W^{(l)}\Omega_m^{\text{in}}c_{k,l+1}^{\text{out}}\Omega_{k,l+1}^{\text{out}}W^{(l+2)}}\right) \\&+
    \frac{\mathcal{U}_L}{W^{(l)}}\left(\left(\Omega_j^{\text{out}}r_j -r_j^2\right)\frac{1}{\Omega_j^{\text{in}}}c_j^{\text{out}}W^{(l+1)}\sum_i{q_{ij}c_{i,l-1}^{\text{in}}\Omega_{i,l-1}^{\text{in}}W^{(l-1)}} - r_j\sum_{i,m\neq j}{q_{im}c_{i,l-1}^{\text{in}}\Omega_{i,l-1}^{\text{in}}W^{(l-1)}c_m^{\text{out}}W^{(l+1)}\Omega_m^{\text{out}}}\right) \mpt
\end{align}
Upon defining the following prefactors,
\begin{align}
    c_j^R&\equiv \sqrt{\frac{\mathcal{U}_R}{W^{(l+1)}}}W^{(l)}c_j^{\text{in}}\mcm \\
    c_{jk}^{(l+1)}&\equiv \sqrt{\frac{\mathcal{U}_R}{W^{(l+1)}}}W^{(l+2)}q_{jk}c_{k,l+1}^{\text{out}}\mcm \\
    c_j^L&\equiv \sqrt{\frac{\mathcal{U}_L}{W^{(l)}}}W^{(l+1)}c_j^{\text{out}}\mcm \\
    c_{ij}^{(l-1)}&\equiv \sqrt{\frac{\mathcal{U}_L}{W^{(l)}}}W^{(l-1)}q_{ij}c_{i,l-1}^{\text{in}}\mcm
\end{align}
we arrive at our final expression for the connectivity dynamics with explicit coupling to adjacent layers,
\begin{align}
    \dv{r_j}{t} &\approx \left((r_j - r_j\sqrt{r_j})c_j^R\sum_k{c_{jk}^{(l+1)}\sqrt{r_k^{(l+1)}}} - r_j\sum_{m\neq j,k}{c_m^Rc_{mk}^{(l+1)}\sqrt{r_m}\sqrt{r_k^{(l+1)}}}\right) \\ &+ \left((r_j - r_j\sqrt{r_j})c_j^L\sum_i{c_{ij}^{(l-1)}\sqrt{r_i^{(l-1)}}} - r_j\sum_{i,m\neq j}{c_m^Lc_{im}^{(l-1)}\sqrt{r_m}\sqrt{r_i^{(l-1)}}}\right) \\
    &= r_j(1 - \sqrt{r_j})\left(c_j^R\sum_k{c_{jk}^{(l+1)}\sqrt{r_k^{(l+1)}}} + c_j^L\sum_i{c_{ij}^{(l-1)}\sqrt{r_i^{(l-1)}}}\right) \\&- r_j\sum_{m\neq j}\sqrt{r_m}\left(\sum_{k}{c_m^Rc_{mk}^{(l+1)}\sqrt{r_k^{(l+1)}}} + \sum_{i}{c_m^Lc_{im}^{(l-1)}\sqrt{r_i^{(l-1)}}}\right)\mpt
    \label{eq:r_dynamics_with_coupling}
\end{align}
Structurally, this equation is very similar to the case without coupling, except we now see that the $c_j$ from before have a dependency on the adjacent layer connectivities.

\subsection{Determining the sign of the growth rate constants}\label{sec:S_growing_weights}
In the derivation above we make the simplification that some of the $\U{s}{t}$ terms have the same magnitude and sign, independent of their arguments,
\begin{equation}\label{eq:U_approximation}
    \U{s}{t} \approx\abs{\U{}{}} \; \forall\, \set{s,t}\mpt
\end{equation}
This is an important assumption, underpinning the results that followed, hence we clarify its origin. The intuition that leads us to this approximation, is the initial state of the network. Let us first recall that in the definition of $\U{s}{t}$, Eq.~\eqref{eq:U_definition}, we sum over all active paths into node $s$ and out of node $t$, contained in the set $\mathcal{A}_{s}^{t}$. In the initial random state of the network, each path is equally likely to be active and contribute to the output, so that we can set the size of $\mathcal{A}_{s}^{t}$, and thus the number of terms in the summation, to be constant and independent of $s$ and $t$. Each summand contains a weight product of an individual path, and the value of this product can fluctuate. However, initially these values are randomly distributed over the pairs $(s,t)$, such that after summing over all paths, these fluctuations cancel, as we have set the number of paths to be equal. We then conclude that the magnitude of $\U{s}{t}$ is constant,
\begin{equation}
    \U{s}{t} \approx \pm\abs{\U{}{}} \; \forall\, \set{s,t}\mpt
\end{equation}
The second assumption regarding the equivalence of the sign is more intricate. Its motivation resides in the fact that these $\U{s}{t}$ terms capture almost the entire weight update, except for the nearest neighbour part. If we start from very small initial weights, then on average each weight will have to grow in order to produce an output of reasonable size. To see this, consider a DNN with $H=10$ and $n_l=10\;\forall\,l\neq H$, $n_H=1$. Let each random initial weight in the network have a magnitude of order $\abs{w}\sim\mathcal{O}\left(10^{-2}\right)$ and assume that the input instances have features that are of order $X_{i}^{(m)}\sim\mathcal{O}\left(1\right)$. Then the total output of the network after initialisation is at most\footnote{Assuming that all paths are active and contribute to the output.} of order
\begin{equation}
    \hat{Y}^{(m)} \sim \mathcal{O}\left(10^{10} \cdot \left(10^{-2}\right)^{10}\right)=\mathcal{O}\left(10^{-10}\right)\mcm
\end{equation}
which follows from multiplying the total number of paths with the weight and feature product of each path. Producing an output that is of order $\hat{Y}^{(m)} \sim \mathcal{O}\left(1\right)$ requires weights to grow significantly in magnitude. Combining this with the fact that a large negative weight is more likely to make its connected node inactive and hence not contributing to the output, we conclude that on average, weights will grow in positive direction, such that we can approximate the sign of all $\U{s}{t}$ terms as being positive. However, we emphasise that this depends on the initial order of magnitude of the weight initialisation. With this, at initialisation and shortly after training has commenced, we arrive at the approximation in Eq.~\eqref{eq:U_approximation}, which is used to derive Eq.~\eqref{eq:coarse_grained_r_dynamics} and Eq.~\eqref{eq:r_dynamics_with_coupling}. A consequence of the assumption of growing weights, is also that all connectivities are trying to grow, regardless of the suppression by other connectivities. Therefore, the `growth rates' $c$ are all assumed to be positive.

\section{Analysis of weight morphologies}
We now want to use the coarse-grained nodal connectivity equations we found above to study the formation of weight structures on larger scales. To this end we first find the homogeneous state of the connectivity dynamics and then perturb this state to see what kind of instabilities arise.

\subsection{Homogeneous state without explicit interlayer coupling}\label{sec:S_hom_state_without_coupling}
In the case without explicit interlayer coupling, the initial homogeneous stable state is given by $r_j^{\text{hom}}=\frac{1}{N^2}$ and $c_j=c\forall j$:
\begin{align}
    \dv{r_j^{\text{hom}}}{t}&= \frac{1}{N^2}(1-\sqrt{\frac{1}{N^2}})c - \frac{1}{N^2}\sum_{m\neq j}{\sqrt{\frac{1}{N^2}}c} \\
    &=\frac{1}{N^2}(1-\frac{1}{N})c - \frac{1}{N^2}(N-1){\frac{1}{N}c} = 0.
\end{align}
This state corresponds to the case where weights are distributed across nodes in such a way that there are no fluctuations on the level of nodal connectivities.

\subsection{Homogeneous state with explicit interlayer coupling}
With coupling, the homogeneous state $r_j^{\text{hom}}=\frac{1}{N^2}$, $c_j^R=c^R$, $c_j^L=c^L$, $c_{jk}^{(l+1)}=c_k^{(l+1)}$ and $c_{ij}^{(l+1)}=c_i^{(l+1)}$ $\forall j$ is stable,
\begin{align}
    \dv{r_j^{\text{hom}}}{t}&=\frac{1}{N^2}(1 - \sqrt{\frac{1}{N^2}})\left(c^R\sum_k{c_{k}^{(l+1)}\sqrt{\frac{1}{N^2}}} + c^L\sum_i{c_{i}^{(l-1)}\sqrt{\frac{1}{N^2}}}\right) \\&- \frac{1}{N^2}\sum_{m\neq j}\sqrt{\frac{1}{N^2}}\left(\sum_{k}{c^R c_{k}^{(l+1)}\sqrt{\frac{1}{N^2}}} + \sum_{i}{c^L c_{i}^{(l-1)}\sqrt{\frac{1}{N^2}}}\right) \\
    &= \frac{1}{N^2}(1 - \frac{1}{N})\left(c^R\sum_k{c_{k}^{(l+1)} \frac{1}{N}} + c^L\sum_i{c_{i}^{(l-1)} \frac{1}{N}}\right) \\&- \frac{1}{N^2}(N-1) \frac{1}{N}\left(c^R\sum_{k}{ c_{k}^{(l+1)} \frac{1}{N}} + c^L\sum_{i}{ c_{i}^{(l-1)} \frac{1}{N}}\right) \\
    &=0.
\end{align}

\subsection{Channels as a highest order instability of the homogeneous state}
Close to the homogeneous state, we can ignore adjacent layer couplings, since the weight feedback described earlier has not yet lead to any strong correlations between weights in separated layers. We thus perturb the homogeneous state of equation~\eqref{eq:coarse_grained_r_dynamics}, which indeed does not capture interlayer couplings, by setting
\begin{align}
    r_j &\to \frac{1}{N^2}+\delta r_j\mcm \\
    c_j &\to c + \delta c_j\mpt
\end{align}
Now we substitute this in the differential equation and only keep terms up to linear order in the perturbations, leading to
\begin{align}
\dv{\delta r_j}{t} = \frac{1}{N^2}\left(\delta c_j - \langle\delta c\rangle\right) - \frac{1}{2}c\langle\delta r\rangle\mpt
\end{align}
Due to normalisation and correlation of the in- and outgoing weight fractions that define the connectivity, the average perturbation $\langle\delta r\rangle$ must be close to zero, as the gain in connectivity of one node must, by normalisation, come at an equal cost of a loss of connectivity of another node. Therefore a perturbation of the homogeneous state is growing if $\delta c_j > \langle\delta c\rangle$.
Staying closer to the original equation, we can also study when $\dv{r_j}{t}>0$, i.e. when is the connectivity itself growing,
\begin{align}
    \dv{r_j}{t} = r_j(1-\sqrt{r_j})c_j - r_j\sum_{m\neq j}{\sqrt{r_m}c_m} &> 0 \\
    r_j \left(c_j - \sum_m{\sqrt{r_m}c_m}\right) &> 0 \\
    c_j &> \sum_m{\sqrt{r_m}c_m}\mpt
\end{align}
Now before we approximated $\sqrt{r_m}~\Omega_m^{\text{in/out}}$, and since these fractions $f$ are normalised, so is $\sqrt{r_m}$. We can thus interpret this as a probability distribution, such that $r_j$ is growing if
\begin{equation}
    c_j > \langle c_m \rangle_{r_m}\mcm
\end{equation}
which means that a growth rate has to be larger than the weighted average of other nodes in the same layer, for this node to outgrow the others.

\subsection{Separation of channel forming instability and instabilities due to layer couplings}

We now perturb the homogeneous state by letting
\begin{align}
    r_j^{(l)} &\to \frac{1}{N^2} + \delta r_j^{(l)}\mcm \\
    c_j^R &\to c^R + \delta c_j^R \mcm \\
    c_j^L &\to c^L + \delta c_j^L \mcm \\
    c_{jk}^{(l+1)} &\to c_k^{(l+1)} + \delta c_{jk}^{(l+1)}\mcm \\
    c_{kj}^{(l-1)} &\to c_k^{(l-1)} + \delta c_{kj}^{(l-1)}\mpt
\end{align}

Substituting first the perturbation in $r$ into the original dynamics,
\begin{align}
    \dv{r_j^{(l)}}{t} &= r_j\sum_{k}\left(c_j^Rc_{jk}^{(l+1)}\sqrt{r_k^{(l+1)}} + c_j^Lc_{kj}^{(l-1)}\sqrt{r_k^{(l-1)}}\right) \\&- r_j\sum_{m}\sqrt{r_m}\sum_k\left(c_m^Rc_{mk}^{(l+1)}\sqrt{r_k^{(l+1)}} + c_m^Lc_{km}^{(l-1)}\sqrt{r_k^{(l-1)}}\right)
\end{align}
gives, using that $\sqrt{\frac{1}{N^2} + \delta r_j^{(l)}}\approx \frac{1}{N}+\frac{N}{2}\delta r_j^{(l)}$,
\begin{align}
    \dv{\delta r_j^{(l)}}{t} &= \left(\frac{1}{N^2} + \delta r_j^{(l)}\right)\sum_{k}\left(c_j^Rc_{jk}^{(l+1)}\left(\frac{1}{N}+\frac{N}{2}\delta r_k^{(l+1)}\right) + c_j^Lc_{kj}^{(l-1)}\left(\frac{1}{N}+\frac{N}{2}\delta r_k^{(l-1)}\right)\right) \\&- \left(\frac{1}{N^2} + \delta r_j^{(l)}\right)\sum_{m}\left(\frac{1}{N}+\frac{N}{2}\delta r_m^{(l)}\right)\sum_k\left(c_m^Rc_{mk}^{(l+1)}\left(\frac{1}{N}+\frac{N}{2}\delta r_k^{(l+1)}\right) + c_m^Lc_{km}^{(l-1)}\left(\frac{1}{N}+\frac{N}{2}\delta r_k^{(l-1)}\right)\right) \\
    &= \frac{1}{N^2}\sum_{k}\left(c_j^Rc_{jk}^{(l+1)}\left(\frac{1}{N}+\frac{N}{2}\delta r_k^{(l+1)}\right) + c_j^Lc_{kj}^{(l-1)}\left(\frac{1}{N}+\frac{N}{2}\delta r_k^{(l-1)}\right)\right) \\
    &+ \delta r_j^{(l)}\sum_{k}\left(c_j^Rc_{jk}^{(l+1)}\left(\frac{1}{N}+\frac{N}{2}\delta r_k^{(l+1)}\right) + c_j^Lc_{kj}^{(l-1)}\left(\frac{1}{N}+\frac{N}{2}\delta r_k^{(l-1)}\right)\right) \\
    &- \frac{1}{N^2}\sum_{m}\left(\frac{1}{N}+\frac{N}{2}\delta r_m^{(l)}\right)\sum_k\left(c_m^Rc_{mk}^{(l+1)}\left(\frac{1}{N}+\frac{N}{2}\delta r_k^{(l+1)}\right) + c_m^Lc_{km}^{(l-1)}\left(\frac{1}{N}+\frac{N}{2}\delta r_k^{(l-1)}\right)\right) \\
    &- \delta r_j^{(l)}\sum_{m}\left(\frac{1}{N}+\frac{N}{2}\delta r_m^{(l)}\right)\sum_k\left(c_m^Rc_{mk}^{(l+1)}\left(\frac{1}{N}+\frac{N}{2}\delta r_k^{(l+1)}\right) + c_m^Lc_{km}^{(l-1)}\left(\frac{1}{N}+\frac{N}{2}\delta r_k^{(l-1)}\right)\right) \mpt \\
\end{align}
We now simplify this equation by collecting terms up to lowest order in the fluctuations and neglecting all higher order, quadratic and more, terms,
\begin{align}
    &= \frac{1}{N^3}\sum_{k}\left(c_j^Rc_{jk}^{(l+1)} + c_j^Lc_{kj}^{(l-1)}\right) \\
    &+ \frac{1}{2N}\sum_{k}\left(c_j^Rc_{jk}^{(l+1)}\delta r_k^{(l+1)} + c_j^Lc_{kj}^{(l-1)}\delta r_k^{(l-1)}\right) \\
    &+ \frac{1}{N}\delta r_j^{(l)}\sum_{k}\left(c_j^Rc_{jk}^{(l+1)} + c_j^Lc_{kj}^{(l-1)}\right) \\
    &- \frac{1}{N^4}\sum_{m}\sum_k\left(c_m^Rc_{mk}^{(l+1)} + c_m^Lc_{km}^{(l-1)}\right) \\
    &- \frac{1}{2N^2}\sum_{m}\sum_k\left(c_m^Rc_{mk}^{(l+1)}\delta r_k^{(l+1)} + c_m^Lc_{km}^{(l-1)}\delta r_k^{(l-1)}\right) \\
    &- \frac{1}{2N^2}\sum_{m}\delta r_m^{(l)}\sum_k\left(c_m^Rc_{mk}^{(l+1)} + c_m^Lc_{km}^{(l-1)}\right) \\
    &- \frac{1}{N^2}\delta r_j^{(l)}\sum_{m}\sum_k\left(c_m^Rc_{mk}^{(l+1)} + c_m^Lc_{km}^{(l-1)}\right) \mpt \\
\end{align}
We now perturb the constants $c$ nd again only keep linear terms in the perturbation,
\begin{align}
    c_m^Rc_{mk}^{(l+1)} + c_m^Lc_{km}^{(l-1)} &\to \left(c^R + \delta c_m^R\right)\left(c_k^{(l+1)} + \delta c_{mk}^{(l+1)}\right) + \left(c^L + \delta c_m^L\right)\left(c_k^{(l-1)} + \delta c_{km}^{(l-1)}\right) \\
    &\approx c^R c_k^{(l+1)} + c^R \delta c_{mk}^{(l+1)} + \delta c_m^R c_k^{(l+1)} + c^L c_k^{(l-1)} + c^L \delta c_{km}^{(l-1)} + \delta c_m^L c_k^{(l-1)} \\ 
    &\equiv c^R c_k^{(l+1)} + \delta_m\left(c^R c_k^{(l+1)}\right) + c^L c_k^{(l-1)} + \delta_m\left(c^L c_k^{(l-1)}\right)\mpt
\end{align}
Again keeping only linear orders also in the cross-perturbation terms, we get for the full perturbed dynamics,
\begin{align}
    &= \frac{1}{N^3}\sum_{k}\left(c^R c_k^{(l+1)} + \delta_j\left(c^R c_k^{(l+1)}\right) + c^L c_k^{(l-1)} + \delta_j\left(c^L c_k^{(l-1)}\right)\right) \\
    &+ \frac{1}{2N}\sum_{k}\left(c^R c_k^{(l+1)}\delta r_k^{(l+1)} + c^L c_k^{(l-1)}\delta r_k^{(l-1)}\right) \\
    &+ \frac{1}{N}\delta r_j^{(l)}\sum_{k}\left(c^R c_k^{(l+1)} + c^L c_k^{(l-1)}\right) \\
    &- \frac{1}{N^4}\sum_{m}\sum_k\left(c^R c_k^{(l+1)} + \delta_m\left(c^R c_k^{(l+1)}\right) + c^L c_k^{(l-1)} + \delta_m\left(c^L c_k^{(l-1)}\right)\right) \\
    &- \frac{1}{2N^2}\sum_{m}\sum_k\left(c^R c_k^{(l+1)}\delta r_k^{(l+1)} + c^L c_k^{(l-1)}\delta r_k^{(l-1)}\right) \\
    &- \frac{1}{2N^2}\sum_{m}\delta r_m^{(l)}\sum_k\left(c^R c_k^{(l+1)} + c^L c_k^{(l-1)}\right) \\
    &- \frac{1}{N^2}\delta r_j^{(l)}\sum_{m}\sum_k\left(c^R c_k^{(l+1)} + c^L c_k^{(l-1)}\right) \mpt\\
\end{align}
This simplifies to
\begin{align}
    &= \frac{1}{N^3}\sum_{k}\left(\delta_j\left(c^R c_k^{(l+1)}\right) + \delta_j\left(c^L c_k^{(l-1)}\right)\right) \\
    &- \frac{1}{N^4}\sum_{m}\sum_k\left(\delta_m\left(c^R c_k^{(l+1)}\right) + \delta_m\left(c^L c_k^{(l-1)}\right)\right) \\
    &- \frac{1}{2N^2}\sum_{m}\delta r_m^{(l)}\sum_k\left(c^R c_k^{(l+1)} + c^L c_k^{(l-1)}\right) \\
    &\equiv \frac{1}{N^3}\left(\delta_j C - \langle\delta_m C\rangle\right) - \frac{1}{2N}C\langle \delta r^{(l)}\rangle\mcm
\end{align}
where
\begin{align}
    C&\equiv \sum_k\left(c^R c_k^{(l+1)} + c^L c_k^{(l-1)}\right) \\
    \delta_m C &\equiv \sum_k\left(\delta_m\left(c^R c_k^{(l+1)}\right) + \delta_m\left(c^L c_k^{(l-1)}\right)\right) \,.
\end{align}
Again using that $\langle \delta r^{(l)}\rangle$ is close to zero, we find that to linear order, there is no coupling to neighbouring layer connectivities, so the channel formation is indeed a `highest order' effect, and the channel amplitude modulations are a second order effect in the perturbation, which we study below. This shows that channel formation, i.e. an instability within each layer, and oscillations, an instability between different layers, are caused by different orders in the perturbation: we can separate them in time.

\subsection{Channel amplitude modulations induced by layer couplings}\label{sec:S_big_R_dynamics}
\subsubsection{Definition of amplitude variable}
To study modifications of the channel structure, we now first introduce a new variable $R^{(l)}\equiv\sum_n r_n^{(l)}$ which quantifies the channel width or amplitude, and derive its dynamics close to the homogeneous state. In the homogeneous state the channel width is minimal at
\begin{equation}
    R^{(l)}_{\text{hom}} = \sum_i r_i^{\text{hom}} = \sum_i \frac{1}{N^2} = \frac{1}{N}
\end{equation}
and its maximal value is reached when one node has the maximum connectivity of 1, i.e. $R^{(l)}_{\text{max}}=1$. In other words, a large value of $R^{(l)}$ corresponds to a narrow channel width, i.e. a small number of nodes with large connectivities, and vice versa.

\subsubsection{Dynamics of the channel amplitude}
We now derive the dynamics of the amplitude variable $R$ using the known dynamics for $r_i$,
\begin{align}
    \dv{R^{(l)}}{t} &= \sum_j \dv{r_j^{(l)}}{t} \\
    &= \sum_j r_j\sum_{k}\left(c_j^Rc_{jk}^{(l+1)}\sqrt{r_k^{(l+1)}} + c_j^Lc_{kj}^{(l-1)}\sqrt{r_k^{(l-1)}}\right) \\&- \sum_j r_j\sum_{m}\sqrt{r_m}\sum_k\left(c_m^Rc_{mk}^{(l+1)}\sqrt{r_k^{(l+1)}} + c_m^Lc_{km}^{(l-1)}\sqrt{r_k^{(l-1)}}\right) \\
    &= \sum_j r_j\sum_{k}\left(c_j^Rc_{jk}^{(l+1)}\sqrt{r_k^{(l+1)}} + c_j^Lc_{kj}^{(l-1)}\sqrt{r_k^{(l-1)}}\right) \\&- R^{(l)}\sum_{m}\sqrt{r_m}\sum_k\left(c_m^Rc_{mk}^{(l+1)}\sqrt{r_k^{(l+1)}} + c_m^Lc_{km}^{(l-1)}\sqrt{r_k^{(l-1)}}\right) \\
    &= \sum_m\left(r_m^{(l)}-R^{(l)}\sqrt{r_m^{(l)}}\right)\sum_k\left(c_m^Rc_{mk}^{(l+1)}\sqrt{r_k^{(l+1)}} + c_m^Lc_{km}^{(l-1)}\sqrt{r_k^{(l-1)}}\right)\mpt
\end{align}
To continue from here, we expand $r_i^{(l)}\approx \frac{1}{N}R^{(l)}\left(\delta r_i^{(l)}\right) + \delta r_i^{(l)}$, $\sqrt{r_i^{(l)}}\approx \sqrt{\frac{R^{(l)}}{N}}\left(1+\frac{N}{2R^{(l)}}\delta r_i^{(l)}\right)$. Plugging this into the above equation, leaving out the explicit dependence of $R$ on the perturbations of $r_i$ for now, and keeping only linear terms in $\delta r_i$ gives
\begin{align}
     \dv{R^{(l)}}{t} &\approx \frac{R^{(l)}}{N}\left(1-\sqrt{R^{(l)}N}\right)\sum_m\sum_k\left(c_m^Rc_{mk}^{(l+1)}\sqrt{r_k^{(l+1)}} + c_m^Lc_{km}^{(l-1)}\sqrt{r_k^{(l-1)}}\right) \\ &+ \left(1-\frac{1}{2}\sqrt{R^{(l)}N}\right)\sum_m\delta r_m^{(l)}\sum_k\left(c_m^Rc_{mk}^{(l+1)}\sqrt{r_k^{(l+1)}} + c_m^Lc_{km}^{(l-1)}\sqrt{r_k^{(l-1)}}\right) \\
     &\approx \frac{R^{(l)}}{N}\left(1-\sqrt{R^{(l)}N}\right)\sum_m\sum_k\left(c_m^Rc_{mk}^{(l+1)}\sqrt{\frac{R^{(l+1)}}{N}} + c_m^Lc_{km}^{(l-1)}\sqrt{\frac{R^{(l-1)}}{N}}\right) \\
     &+ \frac{R^{(l)}}{N}\left(1-\sqrt{R^{(l)}N}\right)\sum_m\sum_k\left(\frac{1}{2}c_m^Rc_{mk}^{(l+1)}\sqrt{\frac{N}{R^{(l+1)}}}\delta r_k^{(l+1)} + \frac{1}{2}c_m^Lc_{km}^{(l-1)}\sqrt{\frac{N}{R^{(l-1)}}}\delta r_k^{(l-1)}\right) \\
     &+ \left(1-\frac{1}{2}\sqrt{R^{(l)}N}\right)\sum_m\delta r_m^{(l)}\sum_k\left(c_m^Rc_{mk}^{(l+1)}\sqrt{\frac{R^{(l+1)}}{N}} + c_m^Lc_{km}^{(l-1)}\sqrt{\frac{R^{(l-1)}}{N}}\right)\mpt
\end{align}
The sign of the first and highest-order term in this expression is governed by
\begin{align}
    1-\sqrt{R^{(l)}N} &=0\mcm \\
    R^{(l)} &= \frac{1}{N}\mpt
\end{align}
This means that this term is only positive, if $R^{(l)} < \frac{1}{N}$, but that never happens, as $\frac{1}{N}\leq R^{(l)}\leq 1$. Therefore, this term is always negative. Since it couples to the magnitude of the neighbouring layer, a large $R^{(l\pm 1)}$ (narrow channel, high connectivity focusing onto a few nodes) increases the value of this negative term and thereby reduces the growth of $R^{(l)}$.
We can simplify the highest-order term a bit further,
\begin{align}
    &\frac{R^{(l)}}{N}\left(1-\sqrt{R^{(l)}N}\right)\sum_m\sum_k\left(c_m^Rc_{mk}^{(l+1)}\sqrt{\frac{R^{(l+1)}}{N}} + c_m^Lc_{km}^{(l-1)}\sqrt{\frac{R^{(l-1)}}{N}}\right) \\
    &\equiv \frac{R^{(l)}}{N\sqrt{N}}\left(1-\sqrt{R^{(l)}N}\right)\left(c^{\text{right}}\sqrt{R^{(l+1)}} +c^{\text{left}}\sqrt{R^{(l-1)}} \right)\mcm
\end{align}
where we defined
\begin{align}
    c^{\text{right}} &\equiv \sum_m\sum_k c_m^Rc_{mk}^{(l+1)}\mcm \\
    c^{\text{left}} &\equiv \sum_m\sum_k \mpt
    c_m^Lc_{km}^{(l-1)}
\end{align}
Although the terms of order $\mathcal{O}(\delta r)$ can be positive, we find that independent of fluctuations on individual nodes, a repressive interaction of the full amplitude always exists.

By defining the rescaled amplitude variable $a=NR$, we get the equation presented in the main text up to the prefactor of $N^{-2}$,
\begin{align}
     \dv{a^{(l)}}{t} &\approx \frac{a^{(l)}}{N^2}\left(1-\sqrt{a^{(l)}}\right)\left(c^R\sqrt{a^{(l+1)}} + c^L\sqrt{a^{(l-1)}}\right) \\
     &+ \frac{a^{(l)}}{2}\left(1-\sqrt{a^{(l)}}\right)\sum_k\left(\frac{\tilde{c}_k^{\text{R}}}{\sqrt{a^{(l+1)}}}\delta r_k^{(l+1)} + \frac{\tilde{c}_k^{\text{L}}}{\sqrt{a^{(l-1)}}}\delta r_k^{(l-1)}\right) \\
     &+ \left(1-\frac{1}{2}\sqrt{a^{(l)}}\right)\sum_m\delta r_m^{(l)}\left(\overline{c}_m^{\text{R}}\sqrt{a^{(l+1)}} + \overline{c}_m^{\text{L}}\sqrt{a^{(l-1)}}\right)\mcm
\end{align}

where we additionally defined
\begin{align}
    \tilde{c}_k^{\text{R}} &\equiv \sum_m c_m^Rc_{mk}^{(l+1)} \mcm\\
    \tilde{c}_k^{\text{L}} &\equiv \sum_m c_m^Lc_{km}^{(l-1)}\mcm
    \\
    \overline{c}_m^{\text{R}} &\equiv \sum_k c_m^Rc_{mk}^{(l+1)}\mcm
    \\
    \overline{c}_m^{\text{L}} &\equiv \sum_k c_m^Lc_{km}^{(l-1)}\mpt
\end{align}

As explained above, the amplitude $a_l$ is bounded between 1 and $N$, such that the factor $1-\sqrt{a_{l}}$ in the first term is always negative or at most equal to 0. Therefore, this term always leads to a decrease in the value of $a_l$. Since this negative factor is multiplied with the channel amplitudes $a_{l\pm 1}$ in adjacent layers, a large value of the adjacent layer amplitude (narrow channel) means a larger negative value of this term. We thus see that this is an interaction term that represents an inhibition by the channel amplitudes in neighbouring layers, leading to local anticorrelations.

\end{document}